\documentclass[twoside,11pt]{article}
\usepackage{jmlr2e}

\usepackage{fullpage}
\usepackage{xcolor}
\usepackage{hyperref}
\usepackage{bbm}
\usepackage{amsmath,amssymb,amsfonts}
\usepackage{graphicx}
\usepackage{natbib}
\usepackage{url,hyperref,lineno,microtype,subcaption}
\usepackage{xfrac}
\usepackage{comment}
\usepackage{multirow}

\def\R{{\mathbb{R}}}
\def\pr{{\rm Pr}}
\def\E{{\mathbb E}}
\def\X{{\mathcal X}}

\def\A{{\mathcal A}}
\def\H{{\mathcal H}}
\def\F{{\mathcal F}}

\def\Z{{\mathcal Z}}

\def\I{{\mathcal I}}
\def\var{{\mbox{\tt var}}}

\renewcommand{\S}{\ensuremath{\mathcal{S}}}

\title{Streaming Encoding Algorithms for Scalable Hyperdimensional Computing}

\date{\today}

\author{\name Anthony Thomas \email ahthomas@eng.ucsd.edu
       \AND Behnam Khaleghi \email bkhalegh@eng.ucsd.edu
       \AND Gopi Krishna Jha \email gopi.krishna.jha@intel.com
       \AND Sanjoy Dasgupta \email dasgupta@eng.ucsd.edu
       \AND Nageen Himayat \email nageen.himayat@intel.com
       \AND Ravi Iyer \email ravishankar.iyer@intel.com
       \AND Nilesh Jain \email nilesh.jain@intel.com 
       \AND Tajana Rosing \email tajana@eng.ucsd.edu}

\begin{document}
%\onecolumn
%\firstpage{1}

%\title[Hashing and HD Computing]{Hashing for Fast and Scalable Encoding in Hyperdimensional Computing} 

%\author[\firstAuthorLast ]{\Authors} %This field will be automatically populated
%\address{} %This field will be automatically populated
%\correspondance{} %This field will be automatically populated

%\extraAuth{}

\maketitle

\begin{abstract}
\noindent Hyperdimensional computing (HDC) is a paradigm for data representation and learning originating in computational neuroscience. HDC represents data as high-dimensional, low-precision vectors which can be used for a variety of information processing tasks like learning or recall. The mapping to high-dimensional space is a fundamental problem in HDC, and existing methods encounter scalability bottlenecks when the input data itself is high-dimensional. In this work, we explore a family of streaming encoding techniques based on hashing. We show formally that these methods enjoy comparable guarantees on performance for learning applications while being substantially more efficient than existing alternatives. We validate these results experimentally on a popular high-dimensional classification problem and show that our approach easily scales to very large data sets. We develop implementations of our techniques in both FPGA and ``in-memory'' architectures and show they lead to substantial performance improvements over existing methods.
\end{abstract}

\section{Introduction}

\noindent Hyperdimensional computing (HDC) is a neurally-inspired technique for data representation and information processing \citep{kanerva2009hyperdimensional}. From the bird's eye view, HDC works by embedding data from its ambient space $\X$ into a high-dimensional space $\H$ wherein all subsequent information processing is performed. The HD embeddings of data are typically low-precision and well suited to simple, on-line, learning methods like the Perceptron and Winnow algorithms \citep{kleyko2018classification,thomas2020theoretical,imani2017voicehd,DiaoGLVQHD2021}. This makes the technique particularly amenable to implementation in low-power, highly parallel hardware like field-programmable-gate-arrays (FPGAs) \citep{salamat2019f5}, and ``in-memory'' computing architectures \citep{karunaratne2019memory}. Indeed, HDC has received substantial attention from the hardware community in recent years as a low-power alternative to conventional ML techniques for learning tasks like classification and clustering \citep{rahimi2018efficient,imani2019fach,imani2019bric,rahimi2017high}.

The embedding $\phi : \X \rightarrow \H$, typically called the ``encoding function'' in the HDC literature, is an object of fundamental concern. In practice, $\phi$ often takes the form of a random embedding. For instance, to represent a symbol $a$ from some discrete alphabet, one typically generates a codeword $\phi(a)$ by sampling from some distribution over $\H$, say from the uniform distribution over $\{+1,-1\}^{d}$ \citep{rahimi2018efficient,frady2018theory,kanerva2009hyperdimensional}. These encodings can then be combined using simple operations in $\H$ to build more complex structures and affect various information processing tasks like learning or recall. While simple, and remarkably powerful theoretically \citep{thomas2020theoretical}, this approach encounters serious limitations when the dimension of the input data is large, for the simple reason that one must maintain a large codebook, sometimes also called a ``clean-up-memory'' or ``item-memory,'' mapping symbols in the alphabet to their $d$-dimensional HD representations.

This issue is particularly salient in the context of problems defined over high-cardinality categorical data, in which the input data is often high-dimensional and sparse. For instance, data from an online retailer might contain a list of products viewed by a customer along with information about each product. While the number of products viewed by any particular consumer might be small – say on the order of hundreds to thousands – the total universe of products is vast – on the order of tens of millions. Our goal is to devise encoding methods that can represent such data, in a way that is useful for learning, without the need to materialize and store an embedding for each symbol in the alphabet.

\vspace{2mm}
\noindent We here draw inspiration from the literature on streaming algorithms \citep{bloom1970space,cormode2005improved,kane2014sparser}, and explore how techniques based on hashing can be used to efficiently generate encodings ``on-the-fly,'' eliminating the need to construct and store a large codebook. Hash-based streaming techniques easily scale to very large data sets, and yield sparse and binary representations of data. Our key contributions are as follows:
\begin{itemize}
    \itemsep-0.3em
    \item We develop a formal analytic framework that allows one to compare the utility of different encoding techniques for classification tasks formulated on HD representations. Our framework yields simple sufficient conditions on the data and encoding function under which an important class of HD based classification techniques will succeed.
    \item Using this framework, we show that there are simple, computationally efficient streaming analogs, based on hashing, to popular existing encoding techniques for discrete and continuous data. These approaches enjoy similar guarantees in the learning setting, while being substantially more memory-efficient and computationally tractable. 
    \item We formally analyze the tradeoff between the encoding dimension, the number and type of hash-functions, and intrinsic properties of the data like the size of the input domain, and a notion of separability between classes on the original representation of the data.
    \item An empirical evaluation of the proposed techniques on a popular large scale classification problem \citep{criteo} validating our theory, and showing that hash-based encodings offer a comparable (or superior) levels of accuracy to existing HDC techniques while being far more scalable. To our knowledge, this represents, by far, the largest scale of data addressed, to date, for HDC based classification.
    \item We develop an efficient and generic FPGA implementation that accelerates classification using HDC with our encoding methods by more than two orders of magnitude, and offers even better improvements in energy use. It is the first FPGA design of HD that handles both numeric and categorical data and uses logistic regression to estimate the model parameters rather than superposition based techniques.
    \item We propose an in-memory design to accelerate numeric and hash-based categorical encoding, which yields more than three orders of magnitude speedup over CPU. Our design leverages the conventional ReRAM crossbar with a minimal peripheral addendum to support HDC encoding and learning.
\end{itemize}
We anticipate that our work will expand the scope of learning problems that can be addressed using HDC by providing simple and efficient encoding techniques that scale to high-dimensional inputs, which are common in practice. More broadly, our work opens what we believe will be a fruitful line of connection between the hardware focused literature on HDC, and the theoretical literature on streaming algorithms. We hope that our work will encourage further convergence between these two important areas of research. 

\section{Background and Related Work}
\label{sec:background}

In the following section we review the relevant background on HDC and hashing along with related work.

\subsection{Background on Hyperdimensional Computing}

Hyperdimensional computing draws its inspiration from research in the neuroscience community that has established high-dimensional distributed representations as a fundamental data type for neural information processing \citep{babadi2014sparseness,stettler2009representations}. The basic idea of HDC is to map an input $x$, living in some space $\X$ to a high-dimensional, low-precision, representation $\phi(x)$, which resides in some $d$-dimensional space $\H$. All subsequent learning and information processing is performed on these high-dimensional representations. 

Traditionally, the dimension of $\H$ is assumed to be larger than that of $\X$. However, this condition is not necessary, and many HD information processing algorithms will work even if encoding reduces the dimension of the input \citep{thomas2020theoretical}. Indeed, we here focus on the setting that the input is already very high-dimensional. In practice, the particular construction of $\phi$ depends on the type of input data, and the constraints placed on the HD space $\H$. Our subsequent discussion largely consists of comparing different encoding methods, so we defer a detailed description of popular techniques for the time being. 

The HD representations of data can be combined to form more complex objects, for instance representing structured data or sequences, using operators for addition and multiplication--typically referred to as ``bundling'' and ``binding'' in the HD literature \citep{kanerva2009hyperdimensional}. The bundling operator, $\oplus : \H\times\H\rightarrow\H$ superimposes two HD points. The binding operator $\otimes : \H\times\H \rightarrow \H$ is used to form tuples, and intuitively, can be thought of as a way to associate an attribute with a value. The bundling and binding operators are commonly implemented using element-wise addition and multiplication, although other choices are possible as well \citep{kanerva2009hyperdimensional,plate2003holographic}. For instance, bundling is sometimes implemented using a thresholded sum \citep{kleyko2018classification}. These structures can typically be \emph{decoded} to recover--at least approximately--the original data, although this is not necessary depending on the application. For a detailed analysis of methods for encoding and decoding different forms of data, the interested reader is referred to \citep{thomas2020theoretical,frady2018theory}.

The HD representations of data can be used for a variety of information processing tasks like classification, clustering, and recall (e.g. memory). Our focus here is on classification. In general, HDC imposes no particular constraints on the choice of learning algorithm and a wide range of techniques have been used in practice. These range from very simple techniques based on centroids/prototypes \citep{kleyko2018classification,rahimi2018efficient,salamat2019f5} to more sophisticated approaches using trainable multi-layer neural networks or other optimization based techniques \citep{frady2021variable,alonso2021hyperembed,imani2017voicehd}. For a recent survey of learning techniques used in HDC, see \citep{kleyko2021survey,KleykoSurveyVSA2021Part2}. In this work, we attempt to strike a balance between these extremes, and focus on classifiers that can be represented as affine functions in HD space, which covers simple prototype based methods as well as more sophisticated techniques like percpetrons \citep{rosenblatt.58}, winnow \citep{littlestone1988learning}, support vector machines \citep{vapnik1998support}, and logistic regression \citep{hastie2009elements}. 

\subsection{Hashing}

Hashing is a fundamental tool in computer science for constructing efficient representations of data. For a good survey of the technical details of hashing, the reader is referred to \citep{vadhan2012pseudorandomness}. Here, we will simply think of a \emph{hash-function} as a map $\psi$ that takes in a member of some set $\A = \{a_{1},...,a_{m}\}$, and returns a non-negative integer in some pre-defined range $[d] = 1 ,..., d$. Moreover, we typically want the output of $\psi$ to appear random in the sense that $\psi(a_{1}),...,\psi(a_{m})$ simulates i.i.d. draws from the uniform distribution over $[d]$. One typically views $\psi$ being drawn randomly from a family of functions $\Psi = \{\psi : \A \to [d]\}$. For instance, one might take $\Psi$ to be a parametric family, and then instantiate a particular $\psi$ via a random draw of parameters. However, in general, there is a tension between the complexity of storing and evaluating $\psi$, and the randomness of its output. This trade-off is commonly formulated using the notion of a $p$-independent family, which may be defined as follows \citep[Definition 3.31]{vadhan2012pseudorandomness}:

\begin{definition}
\label{defn:independent-family} \textbf{$p$-Independent Family} Let $\A$ be an alphabet of $m$ symbols. A family of hash-functions $\Psi = \{\psi : \A \to [d]\}$ is said to be $p$-independent if, for all sets $\S = \{a_{1},...,a_{p}\} \subset \A$ of size $p < m$, and any $\psi$ chosen uniformly at random from $\Psi$, the random variables $\psi(a_{1}),...,\psi(a_{p})$ are mutually independent and uniformly distributed in $[d]$.
\end{definition}

In general, stronger independence assumptions (bigger $p$) allow one to prove stronger guarantees about the performance of hash-based algorithms. It can be shown that $O(p \max(\log m, \log d))$ bits are sufficient to describe a $p$-independent hash-function \citep{vadhan2012pseudorandomness}. In our our setting, $d \ll m$, and so this can be simplified to $p \log m$. We will return to the question of how much independence is required of our hash-functions in greater detail in Section \ref{sec:hash-encoding}.

\subsubsection{Hashing for Learning Applications and HDC}

There is a long history of using hashing for efficient data representation in the broader machine learning literature \citep{bloom1970space,broder1997resemblance,cormode2005improved,chen2015compressing,shi2009hash}. These techniques have evolved to become a well established part of the empirical toolkit in machine learning \citep{abadi2016tensorflow,meng2016mllib}. Much like the encoding function of HDC, the basic idea is to map a high-dimensional input, into a lower-dimensional representation that preserves the essential information for a particular task. The encoding techniques we pursue here are in this tradition. Indeed, one of our main contributions is to show how the Bloom filter \citep{bloom1970space}, a canonical hash-based method for representing sets, can be used as an efficient encoding strategy with provable guarantees for learning in HDC.

To the best of our knowledge, these techniques have not been explored in any detail in the literature on HDC. Work in \citep{kleyko2019autoscaling,thomas2020theoretical} observed, as we do here, that Bloom filters can be viewed as an HD architecture in which the codewords are sparse and binary, and the bundling operator is the element-wise logical or. The work of \citep{kleyko2020autoscaling} also gives an interesting extension to the basic Bloom-filter scheme that can dynamically adjust its capacity. However, the analysis of these works has focused on quantifying their capacity to store and recall specific data items, which is a distinct problem from using them as input to classification procedures as we consider here. In particular, they analyze a specific decoding scheme that is not germane to our setting, and it is not clear to us that, because one can decode a particular representation, that it is also suitable as an input for learning.

Along with our hardware implementations, one of our main contributions is to develop the theoretical foundations of hash-based encoding methods for use in HD learning algorithms. We give sufficient conditions under which hash-based encoding will succeed for a broad family learning applications, and formally analyze the tradeoff between the encoding dimension, the number of hash-functions, and intrinsic properties of the data like the size of the categorical alphabet, and a notion of separability between classes on the original version of the data.

Our work is also related to literature on methods for constructing \emph{sparse} encodings in HDC \citep{kanerva1988sparse,rachkovskij2001binding,laiho2015high,frady2021variable}. Indeed, some of this work \citep{rachkovskij1990audio,rachkovskij2001binding} uses the element-wise or as a bundling operator, and is also related to Bloom filters. In particular, \citep{rachkovskij2001binding} present a very interesting extension of this scheme that allows one to represent more complex forms of structured data via a technique known as ``context-dependent-thinning.''

Sparse representations are advantageous because one may store them as a list of $(\texttt{index},\texttt{value})$ pairs, possibly reducing memory use. However, explicitly storing a map between symbols and their sparse-encodings still entails linear scaling with the alphabet size, leading to memory bottlenecks. Hashing based approaches can be seen as a method for constructing sparse encodings on the fly, which has the advantage that one does not need to store a lookup table, mapping a symbol to its (sparse) HD representation. Moreover, to our knowledge, existing formal work has focused primarily on encoding and decoding data (e.g. memory), whereas our focus is on learning. We provide new theoretical insight into this setting, which clarifies the relative strengths and weaknesses of sparse and dense encoding methods--in particular the memory savings of sparse encoding methods--along with a detailed empirical evaluation on a learning problem.

\subsection{Related Work on Hardware Acceleration}

HDC has received significant attention from the hardware community.
Computational operations can typically be carried out independently, leading to vector-long parallelism, and the operations are often either binary or low-precision, requiring inexpensive compute blocks.
In this context, FPGA-based implementations of HD stand out \citep{salamat2019f5, manuel2019hardware, imani2021revisiting}, as these devices generally consume less power than other parallel computing platforms such as GPUs.
Moreover, FPGAs are configurable, and can be adapted to suit the complexity of target operations. For instance, to support simple operations such as binary addition or low-bit multiplication, FPGAs can allocate fewer resources and end up with more effective resource utilization.
Critically, for our purposes, the design flow of FPGAs can fuse and pipeline operations; hence, composite operations--like hashing--can be effectively carried out in one cycle.
Nevertheless, most FPGA implementations have targeted applications with numeric data \citep{ge2020classification}, and, to our knowledge, none of them deal with large alphabet sizes that require storing or generating even thousands of vectors.
Most similar work to ours is \citep{manuel2019hardware} which quantizes numeric data into $q$ bins and materializes the vector of each bin on the fly by passing the data through a connectivity logic matrix of $m \times d$ (where $d$ is the HD-dimension).
As the matrix grows with alphabet size $m$, this approach is not sustainable for large alphabets.
A related approach is to use permutation to generate the other vectors from a seed vector \citep{khaleghi2022generic}.
While this technique can be used for categorical data, our evaluation in Section \ref{sec:empirical-evaluation} reveals the deficient performance of this approach due to overhead caused by the data movement.

Another body of research has leveraged the memory-centric nature of HD for in-memory implementation.
Most of these studies use the search capability of the memory crossbar for fast similarity computation between a pre-encoded query and a set of stored prototypes representing classes \citep{imani2017exploring, kazemi2021mimhd, wu2018hyperdimensional}.
Work in \citep{karunaratne2019memory}, however, also stores the alphabet vectors in a codebook and performs simple $n$-gram encoding by reading the alphabet vectors of consecutive feature windows of size $n \simeq 3$.
Such a design is not scalable for applications with a large input domain.
Moreover, their architecture supports bundling encoding with a limited ($\leq n$) shift of the vectors, which is useful for use-cases such as text language detection, but cannot implement applications with numeric data or inputs wherein the spatio-temporal positions of features matter.

Unlike previous works, our fully-parameterized FPGA implementation supports numeric and categorical data (with hash-encoding), virtually unlimited alphabet size, and a user-specifiable number of features and embedding-dimension.
Additionally, for learning, our implementation, for the first time, uses logistic regression and estimates the model parameters using mini-batch stochastic gradient descent (SGD) rather than simple bundling.
The user can also specify the precision of the operands, e.g., numeric features.
Hence, our implementation can be reused for a wide variety of different problems and FPGA sizes by changing the parallelism degree and vector length.
Our in-memory architecture builds upon a generic memory crossbar (hence it can also accelerate other applications such as neural networks) and supports encoding of both categorical and numeric data rather than just similarity search.

\section{Problem Formulation and Data Model}
\label{sec:data-model}

We here focus on data with the following characteristics:
\begin{enumerate}
    \itemsep-0.1em
    \item The data contains a mix of categorical (e.g. discrete) and numeric features.
    \item The data is gathered continuously in a streaming fashion.
    \item The categorical features are drawn from a large alphabet, say on the order of tens-to-hundreds of millions of symbols, which may not be known in advance.
\end{enumerate}
Let $x_{n} \in \X \subset \R^{n}$, be a vector of $n$ ``numeric'' features which lie in some Euclidean space, and let $x_{c}$ be a vector of $s$ categorical features. We assume that each coordinate $x_{c}^{(i)}$ is drawn from some discrete alphabet $\A^{(i)}$, and, without loss of generality, that $\A^{(i)}\cap\A^{(j)} = \varnothing$. Let $\A = \cup_{i}\, \A^{(i)}$, and let $|\A| = m$. In other words, we may think of each $x_{c}$ as a set of $s$ items drawn from $\A$. Let $b(x_{c}) \in \{0,1\}^{m}$ denote the $s$-hot encoding of $x_{c}$. That is, $b(x_{c})$ is an $m$-dimensional vector with exactly $s$ non-zero values, whose positions encode the identity of the symbols in $x_{c}$.

We assume the classification task is binary, and can be modeled as:
\[
    y = f(x_{n},x_{c}) \geq 0, \text{ where } f(x_{n}, x_{c}) = \theta_{n} \cdot x_{n} + \theta_{c} \cdot b(x_{c}) + \nu
\]
where $y \in \pm 1$, $\theta_{n}, \theta_{c}$ are real-valued parameter vectors, and $\nu \in \R$ is an intercept. The assumption of a binary classification task is primarily for clarity of exposition, and our results can be extended to support multi-class problems via techniques like ``one-versus-rest'' decision rules. Our goal is to devise an encoding procedure $\phi : \X \times \{0,1\}^{m} \rightarrow \H$, such that, for all $(x_{n}, x_{c})$:
\[
    \theta \cdot \phi(x_{n},x_{c}) + \nu' \geq 0 \Leftrightarrow \theta_{n} \cdot x_{n} + \theta_{c} \cdot b(x_{c}) + \nu \geq 0,
\]
for some vector of parameters $\theta \in \R^{d}$. In general, for this property to be satisfied, it is sufficient for the encoding to preserve dot-products in the following sense:
\begin{definition}
\label{defn:distance-preservation}
\noindent \textbf{$\Delta(d)$-Dot-Product Preserving Encoding}. Let $\X,\H$ be inner-product spaces of dimension $m$ and $d$ respectively. Let $\phi$ be an encoding function from $\X \rightarrow \H$. We say $\phi$ is dot-product preserving if, for all $x,x' \in \X$:
\[
    x \cdot x' - \Delta(d) \leq \phi(x)\cdot\phi(x') \leq x\cdot x' + \Delta(d),
\]
where $\Delta(d)$ is a noise term that depends on $d$.
\end{definition}
Intuitively, randomness in the encoding process adds noise, which generally can be made small by increasing the encoding dimension ($d$). The precise form of $\Delta(d)$ depends on the encoding method in question. One our main objectives will be to analyze this quantity for different types of encoding. 

One should expect that the choice of $d$ is related to how well separated the data is to begin with: if the data is well separated, then more noise can be tolerated and so $d$ can be chosen smaller, which is desirable computationally. We make this intuition precise in the following theorem. We first remind the reader that the \emph{convex hull} of a set $\X$, denoted $\texttt{conv}(\X)$, is the smallest convex set containing $\X$, or equivalently, the set of all possible convex combinations of points in $\X$.
\begin{theorem}
\label{thm:separability}
Let $\phi : \R^{m} \to \H \subset \R^{d}$ be a $\Delta(d)$-dot-product preserving encoding, and let $\Z,\Z' \subset \R^{m}$ be two sets of points satisfying:
\[
    \gamma = \|p - q\|_{2}^{2} > 0,
\]
where $p,q$ are the closest pair of points in $\texttt{conv}(\Z)$ and $\texttt{conv}(\Z')$ respectively.
Then, there exists $\theta \in \H$ and $\nu \in \R$, such that:
\[
    \theta \cdot \phi(x) + \nu > 0 \text{ for all } x \in \Z, \text{ and } \theta \cdot \phi(x') + \nu < 0, \text{ for all } x' \in \Z',
\]
provided $\Delta(d) < \gamma / 6$.
\end{theorem}
A proof can be found in Appendix \ref{app:proof-thm-separability}. The parameter $\gamma$ quantifies how well separated the two sets are to begin with. If $\gamma$ is large, then we can tolerate more distortion in the encoding, which generally means that $d$ can be smaller. We note that $\theta$ can be written explicitly in the form:
\[
    \theta = \sum_{i=1}^{m+1} \alpha_{i}\phi(x_{i}) - \sum_{i=1}^{m+1} \beta_{i}\phi(x_{i}') = \phi(p) - \phi(q),
\]
for some set of points $\{x_{i}\}_{i=1}^{m+1} \subset \Z$, and $\{x_{i}'\}_{i=1}^{m+1} \subset \Z'$ and associated weights $\alpha_{i},\beta_{i} \in \R$. Thus, $\theta$ can be seen as a generalization of the standard prototype based classifiers commonly used in HDC.

\begin{remark}
\label{remark:pointwise}
The previous theorem is quite strong and guarantees that every possible input point will be correctly classified. However, this requires that $\Delta(d)$-preservation be satisfied for all possible pairs of input points, which may require $d$ to be large. However, we can weaken the theorem to only imply that any arbitrary point is correctly classified, with high-probability, in exchange for a looser condition on $\Delta(d)$. Specifically, this only requires $\Delta(d)$-preservation to hold between an arbitrary query point $x_{o}$, and a single fixed set of points of size at most $2m + 2$. That is to say, for any arbitrary $x_{o}$, it is the case that $x_{o} \cdot x_{i} - \Delta(d) \leq \phi(x_{o})\cdot\phi(x_{i}) \leq  x_{o} \cdot x_{i} + \Delta(d)$, for a fixed set of points $\Z = \{x_{1},...,x_{2m+2}\}$. In this case, we say $\phi$ is $\Delta(d)$-dot-product preserving with respect to $\Z$.
\end{remark}

In the remainder, we will explore specific constructions of $\phi$ that satisfy Definition \ref{defn:distance-preservation}, for both categorical and numeric data and characterize their relative strengths and weaknesses.

\section{Encoding High-Dimensional Categorical Data}
\label{sec:hash-encoding}

In the following section, we compare methods for encoding high-dimensional categorical data. We first address the standard method based on generating encodings by random sampling, and then turn to hash-based methods which address some shortcomings of this approach.

\subsection{Generating Dense Codes by Sampling}

Let $x_{c} = \{a_{1},...,a_{s}\}$ be a categorical feature vector with $s$ components. The conventional approach to encoding such data in HDC is to simply assign each symbol in the alphabet a encoding generated by random sampling from some distribution over $\H$ \citep{kanerva2009hyperdimensional,frady2018theory,thomas2020theoretical}. For instance, one might choose: $\phi(a) \sim \text{Unif}(\{+1,-1\}^{d})$, for all $a \in \A$. That is to say, each coordinate is sampled independently from the uniform distribution over $\{+1,-1\}$. To generate the encoding for a feature vector, we simply bundle together the encodings for each constituent symbol:
\begin{gather}
    \label{eqn:bundle}
    \phi(x_{c}) = \bigoplus_{a \in x_{c}} \phi(a).
\end{gather}
In the following, let us take $\oplus$ to be the element-wise sum. Then, it can be shown that this scheme satisfies Definition \ref{defn:distance-preservation} and, hence, is separability-preserving with respect to the binary encoding $b(x_{c}) \in \{0,1\}^{m}$:
\begin{theorem}
\label{thm:dense-codes}
Let $\phi(a) \sim \text{Unif}(\{\pm 1\}^{d})$ for all $a \in \A$, and let $\phi(x_{c})$ be as defined in Equation \ref{eqn:bundle}, where the bundling operator is the element-wise sum. Then, with probability at least $1-\delta$:
\[
    b(x_{c})\cdot b(x_{c}') -  4\sqrt{\frac{2s^{3}}{d}\log\frac{m}{\delta}} \leq \frac{1}{d}\phi(x_{c})\cdot\phi(x_{c}') \leq b(x_{c})\cdot b(x_{c}') + 4\sqrt{\frac{2s^{3}}{d}\log\frac{m}{\delta}},
\]
for all $x_{c},x_{c}'$.
\end{theorem}
A proof can be found in Appendix \ref{app:proof-dense-codes}. The bound suggests to take $d = O((s^{3}\log m)/\gamma^{2})$ to achieve the strong, uniform version of Theorem \ref{thm:separability}, and $d = O((s^{2}\log m)/\gamma^{2})$ to achieve the weaker pointwise version. We make the following remarks concerning the setup above:

\begin{remark}
In one sense, this is already a substantial improvement over working with the original version of the data. Since $d$ scales with the logarithm of $m$, we have reduced the number of parameters to estimate from $m$ to $\approx s^{2} \log m$. To emphasize the practical significance, in the context of the classification task discussed in Section \ref{sec:empirical-evaluation}, $m \approx 10^7, s = 26$, and $s^{2} \log m \approx 10,900$ (neglecting the effect of the margin, which may be substantial in practice). However, this comes at the expense of storing a codebook for each of the $m$ symbols, which may be infeasible when $m$ is large. Thus, we seek alternative approaches that allow us to compute encodings ``on-the-fly,'' without needing to store a codebook.
\end{remark}

\begin{remark}
One approach to generating encodings on-the-fly, which is already well known in the HDC literature, is based on permutation \citep{kanerva2009hyperdimensional,rahimi2017hyperdimensional,kim2018efficient}. Let $\pi$ be a permutation on $[d]$ with cycle time at least $m$. Then, as before, one generates a encoding $\phi(a_{1})$ using random sampling. However, subsequent encodings are generated by permuting this base-encoding--e.g. $\phi(a_{2}) = \pi(\phi(a_{1}))$, $\phi(a_{3}) = \pi(\pi(\phi(a_{1}))) = \pi^{2}(\phi(a_{1}))$, and so on. While this approach does not require one to explicitly materialize a codebook, it rapidly becomes bottlenecked by excessive computation and data movement as $m$ grows large. In practice, the permutation is often taken to be a cyclic shift, which can be implemented far more efficiently than a generic permutation. However, this imposes the restriction that $d = O(m)$, else one cannot possibly represent an alphabet of size $m$, and hence loses the benefit of reducing the number of parameters to be estimated.
\end{remark}

\subsection{Hashing Methods for Encoding Categorical Data}

We now turn to our main contribution, which is to develop hashing based methods for encoding categorical data. As we will see, hashing provides a mechanism to essentially replicate the functionality of the random coding scheme described above, but without the need to materialize a large codebook.

\subsubsection{Generating Dense Codes by Hashing}

A trivial approach is to simply use hash-functions as a mechanism for generating dense encodings on the fly. Let $\psi_{1},...,\psi_{d}$ be a set of $d$ independent hash-functions from $\A \rightarrow [\pm 1]$. Then we merely define $\phi(a)$ coordinate-wise by $\phi(a)_{i} = \psi_{i}(a)$. Assuming $\psi$ really does simulate independent draws from $\{+1,-1\}$, this is equivalent to sampling $\phi(a) \sim \text{unif}(\{+1,-1\}^{d})$. Encoding then proceeds as usual, by bundling together the encodings for each symbol in the feature vector in question. However, each embedding requires evaluating a relatively large number of hash-functions, which is computationally burdensome. In the following section, we explore \emph{sparse} variants of the scheme described above, that require evaluating far fewer hash-functions.

\subsubsection{Generating Sparse Codes by Hashing}

A natural approach to speed up the embedding described in the previous section is to make the encodings for each symbol \emph{sparse}. That is, to consider encodings where only $k \ll d$ coordinates are non-zero. Our approach is based on the ``Bloom filter,'' a canonical approximate data structure for representing sets \citep{bloom1970space}. The Bloom filter is advantageous in that the encodings, embeddings, and intermediates are fully binary, in contrast to the dense schemes described above. Moreover, the representations produced by a Bloom filter are often \emph{sparse} which can simplify subsequent computation. For instance, the dot-product used to perform inference---$\theta_{c} \cdot \phi(x_{c})$---simplifies in this case to a lookup, followed by a sum, eliminating any multiplications.

Let $\psi_{1},...,\psi_{k}$ be independent hash-functions from $\mathcal{A} \rightarrow [d]$, where $[d]$ denotes the integers $1,2,...,d$, and define the encoding $\phi : \A \rightarrow \{0,1\}^{d}$ coordinate-wise by:
\begin{gather}
    \label{eqn:bloom-encodings}
    \phi(a)_{i} = \begin{cases} 1 &\text{ if any } \psi_{j}(a) = i \\ 0 &\text{ otherwise.} \end{cases}
\end{gather}
The embedding of a feature vector is defined, coordinate-wise, as:
\begin{gather}
    \label{eqn:bloom-encoding}
    \phi(x_{c})_{i} = \underset{a \in x_{c}}{\max}\,\phi(a)_{i}.
\end{gather}
The Bloom filter can be said to \emph{represent} a set in the sense that it supports (approximate) membership queries. To query for membership, we simply use a thresholded dot-product. That is, we say $a \in x_{c}$ if $\phi(a) \cdot \phi(x_{c}) \geq k$ and $a \notin x_{c}$ otherwise \citep{broder2004network}. As has been noted previously, similar architectures have been the focus of prior work in the broader HD literature \citep{rachkovskij1990audio,rachkovskij2001binding,kleyko2019autoscaling}. However, the analysis of which we are aware focuses primarily on the decoding problem (e.g. testing for membership). We here provide analysis directly substantiating their use as input for learning algorithms.

Critically, for our purposes, the dot-product between the Bloom filters representing two sets can be used to estimate the size of their intersection. This is important because it means we do not need to decode the filter to use it for learning. While this general fact is known (see, for instance, \citep{broder2004network}), we are unaware of analysis providing sufficiently tight error bounds for our setting. In particular, the techniques of which we are aware are not tight enough to show that the encoding scheme described in Equation \ref{eqn:bloom-encoding} actually does reduce memory use. We provide such bounds in the following Theorem:
\begin{theorem}
\label{thm:bloom-separability}
Let $x_{c}, x_{c}'$ be sets, each of size $s$, drawn from an alphabet $\A$ of size $m$. Let $\phi(x),\phi(x')$ be as defined in Equation \ref{eqn:bloom-encoding}, where we assume the hash-functions $\psi_{1},...,\psi_{k}$ are drawn uniformly at random from a $2s$-independent family. Then, for all $x_{c},x_{c}'$, with probability at least $1-\delta$:
\[
    \left|\frac{1}{k} \phi(x_{c})\cdot\phi(x_{c}')  - b(x_{c})\cdot b(x_{c}') - \frac{s^{2}k}{2d}\right| \leq \max\left\{ \sqrt{\frac{2s^{3}}{d}\log\frac{m}{\delta}}, \frac{4s}{3k}\log\frac{m}{\delta} \right\}
\]
\end{theorem}
A proof can be found in Appendix \ref{app:proof-bloom-separability}. The theorem implies that it is sufficient to take $k = O((s\log m)/\gamma)$ and $d = O((s^{3}\log m )/\gamma^{2})$ to achieve the uniform version of Theorem \ref{thm:separability}. To achieve the weaker pointwise result, it is sufficient to take $k = O(\log(m) / \gamma)$, and $d = O((s^{2}\log m)/\gamma)$. 

To emphasize: the result is a substantial improvement over the situation in Theorem \ref{thm:dense-codes}. Instead of storing a codebook of size $md \approx (ms^{2}/\gamma^{2})\log m$, we can achieve the same result by simply evaluating $k \approx (\log m )/\gamma$ hash-functions, which can easily be done on-the-fly. Moreover, the encoding rule is simpler, and there is no need to materialize the full embedding at all: one can simply store the indices of the non-zero values. Thus, this method gives us the best of all the previous approaches: like the naive random coding strategy, $d$ scales with $\log m$, which reduces the number of parameters to be estimated, but the encoding operation can be easily computed on-the-fly, and so easily scales to very large alphabets.

\subsubsection{Construction of hash-functions}
\label{subsec:hash-construction}

Of course, the practical usefulness of these methods depends heavily on the particular construction of the hash-functions ($\psi$). An obvious construction would be to simply represent $\psi$ as a table that maps each $a \in \A$ to an integer chosen uniformly at random from $[d]$. However, this requires $O(m \log d)$ memory, and so does not avoid the linear scaling with $m$. Our proof of Theorem \ref{thm:bloom-separability} requires $2s$-wise independence (as introduced in Definition \ref{defn:independent-family}). Storing each such hash-function requires $O(s \log m)$ memory and so the memory requirements of the construction in Theorem \ref{thm:bloom-separability} are $O(sk\log m) = O((s^{2}(\log m)^{2})/\gamma)$ or $O((s(\log m)^{2})/\gamma)$, allowing us to finally eliminate the linear factor in $m$. However, in practice we might hope to use even simpler hash-functions that satisfy just pairwise independence.

This is typically justified by treating the symbols being hashed as draws from a probability distribution over $\A$ \citep{shaltiel2011introduction,vadhan2012pseudorandomness}. Provided this distribution is sufficiently entropic, then the distribution over the output of even just pairwise independent hash-functions will closely approximate an independent draw from the uniform distribution over $[d]$, an observation formalized in a celebrated result called the Leftover Hash lemma \citep{impagliazzo1989how,impagliazzo1989pseudo}. Functions that posess this property are called ``randomness extractors'' \citep{nisan1996randomness}.

Pairwise hash-functions are appealing because they are computationally efficient: even the simplest constructions require just $O(\log(\max(m,d)))$ bits \citep[Theorem 3.26]{vadhan2012pseudorandomness}, allowing us to further reduce memory use by a factor of $s$. The use of pairwise independent hash-functions in Bloom filters was analyzed in \citep{mitzenmacher2008simple} who quantify precisely how much randomness is required of the data distribution. Since our construction using $2s$-wise independence already provides a substantial improvement, we leave a rigorous extension of their analysis to our setting for future work. We include this discussion here to emphasize that one should expect even very simple hash-functions to work well in practice - which we indeed find to be the case in our empirical evaluation.

\section{Methods for Encoding Numeric Data}
\label{sec:numeric-encoding}

We now turn to methods for encoding numeric data. We here focus on a general family of techniques based on \emph{random projection}. There are other ways to encode numeric vector data (see \citep{thomas2020theoretical,kleyko2021survey} for detail), but random projections are very general, and have theoretically appealing properties, and so we focus on them here. As in the case of categorical data (Section \ref{sec:hash-encoding}), we will consider methods yielding both dense and sparse encodings. Let $x_{n} \in \R^{n}$, be a numeric vector to encode. In general, random projection methods take the form:
\[
    \phi(x_{n}) = q(\Phi x_{n}),
\]
where $\Phi \in \R^{d \times n}$ is a matrix whose rows are drawn i.i.d. from some distribution over $\R^{n}$, and $q$ is a non-linearity applied element-wise. Random projection methods are capable of capture very diverse notions of structure in data, and the resulting representations can have powerful properties for learning. For instance, in certain cases, linear separators on random-projection encoded data can capture nonlinear decision boundaries on the original version of the data \citep{rahimi2008random,raginsky2009locality}. We here focus on two instances of this method that are of particular relevance in HDC.

\subsection{Generating Dense Codes by Sampling}

Random projection encoding techniques are generally well known in HDC (see \citep{kleyko2021survey} and the references therein). For completeness, we briefly restate one method that can be used to produce dot-product preserving embeddings. Denote by $\S^{n-1}$ the unit-sphere in $n$-dimensions. Let $x_{n} \in \S^{n-1}$ be a point to encode, and let $\Phi \in \R^{d \times n}$ be a matrix whose rows are sampled from the uniform distribution over $\S^{n-1}$. Now let $\phi(x_{n}) = \text{sign}(\Phi x_{n})$, where $\text{sign}(u)$ returns $+1$ if $u \geq 0$, and $-1$ otherwise. Then, one can show that, to a first order approximation, for any $x_{n},x_{n}' \in \S^{n-1}$ \citep[Corollary 19]{thomas2020theoretical}:
\begin{gather}
    \label{eqn:dense-rp}
    \frac{2}{\pi}(x_{n}\cdot x_{n}') - O(1/\sqrt{d}) \leq \frac{1}{d}\phi(x_{n})\cdot\phi(x_{n}')  \leq \frac{2}{\pi}(x_{n}\cdot x_{n}') + O(1/\sqrt{d}),
\end{gather}
We remark that the codes generated using this method can be regarded as a form of ``locality-sensitive-hash'' in which the collision probability captures a distance function of interest (in this case, the angular distance) on the original data. Note as well that the distortion doesn't depend on the ambient dimension of the data ($n$), which is appealing when $n$ is very large. One sometimes also quantizes $\Phi$ to be low-precision and/or sparse \citep{rachkovskij2015formation,rachkovskij2012randomized,rachkovskij2015estimation,imani2019bric}

\subsection{Generating Dense Codes by Hashing}

As was the case with categorical data, the methods above require one to materialize the embedding matrix $\Phi \in \R^{d \times n}$, which may be infeasible when $n$ (the dimension of the underlying data) is large. To address this issue, we again look to techniques based on hashing. One such approach is the \emph{sparse Johnson-Lindenstrauss transform} (SJLT) \citep{dasgupta2010sparse,kane2014sparser}, which is a procedure for constructing a sparse and low-precision embedding matrix ($\Phi$) that has an efficient implementation using hashing \citep{kane2014sparser}. There are several possible constructions for the SJLT. The following is one option \citep{cohen2018simple}.

Let $k \leq d$ be a positive integer, $x_{n} \in \R^{n}$ be a point to encode, $\sigma$ be a hash-function $[n] \to \{+1,-1\}$, and let $\eta$ be a hash-function $[n] \to [d/k]$, where we assume for simplicity that $d$ is divisible by $k$. Now, let us define an embedding $\phi^{(1)} : \R^{n} \to \R^{(d/k)}$ coordinate-wise by:
\[
    \phi(x_{n})_{i}^{(1)} = \sum_{j=1}^{n}\mathbbm{1}(\eta(j) = i)\sigma(j)(x_{n})_{j} \text{ for } i = 1,...,d/k,
\]
where $\mathbbm{1}(\star)$ is the indicator function that returns $+1$ if its argument is true, and $0$ otherwise. One then simply concatenates $k$ such embeddings, each obtained with a fresh draw of hash-functions, to yield the final embedding, $\phi : \R^{n} \to \R^{d}$:
\begin{gather}
    \label{eqn:sjlt}
    \phi(x) = [\phi(x)^{(1)},...,\phi(x)^{(k)}]^{T}.
\end{gather}
Then, it can be shown that the embeddings preserve pairwise similarities in the sense of Definition \ref{defn:distance-preservation} (see for instance: \citep{kane2014sparser,cohen2018simple}). In general, the resulting embeddings are of high-precision (e.g. real numbers). If this is problematic, one can simply compose these encodings with the dense random-projection technique described in Equation \ref{eqn:dense-rp}, which is efficient since $d \ll n$ after applying the SJLT.

\subsection{Generating Sparse Codes} \label{ref:num-sparse}

In general, sparse and binary encodings are desirable in that they can be stored more efficiently and can often be used to simplify subsequent computation. A popular approach is to sparsify the output of a random-projection either by thresholding \citep{rachkovskij2015formation,rachkovskij2012randomized,rachkovskij2015estimation,dasgupta2020expressivity}, or via a $k$-winner-take-all operation \cite{kanerva1988sparse,babadi2014sparseness,dasgupta2018neural,dasgupta2020expressivity}.

The following approach is analyzed in detail in \citep{dasgupta2018neural,dasgupta2020expressivity}. Let $\Phi \in \R^{d \times n}$, be a matrix whose rows are drawn from $\text{Unif}(\S^{n-1})$, and let $z_{i} = \Phi^{(i)}\cdot x_{n}$, where $\Phi^{(i)}$ is the $i$-th row of $\Phi$. Now define:
\begin{gather}
    \label{eqn:sparse-random-projection}
    \phi(x_{n})_{i} = \begin{cases} 1 &\text{ if $z_{i}$ is in the $k$ largest values of $z$} \\ 0 &\text{ otherwise.}   \end{cases}
\end{gather}
Intuitively, the $\Phi^{(i)}$ define a set of ``receptive-fields'' that are activated for inputs $x_{n}$ that lie within a ball of a certain radius. The precise sense in which this is true is somewhat complex, but intuitively, if $\phi(x_{n})\cdot\phi(x_{n}') = k_{o}$, for $k_{o} \leq k$, then $x_{n},x_{n}'$ must lie in the intersection of the receptive fields of $k_{o}$ different centers, which constrains their maximum distance. The radii in the receptive fields can be tuned by the choice of $d$ and $k$ \citep[Lemma 2]{dasgupta2020expressivity}. Holding $k$ fixed, increasing $d$ reduces the radii of the receptive fields, and hence means that the encoding can differentiate between closer points. 

A practical difficulty with this method is that it requires identifying the $k$-largest coordinates in $z = \Phi x_{n}$, which may be computationally burdensome. However, one can show that a similar locality-preserving property can be satisfied by selecting a threshold $t$ such that, $\Pr(|\Phi^{(i)} \cdot x_{n}| \geq t) = k/d$. We examine this approach empirically in Section \ref{sec:empirical-evaluation}, and show that it offers comparable performance to the dense random projection methods, while enjoying the computational/storage benefits of a sparse binary representation.

\subsection{Methods for Combining Dense and Spare Encodings}
\label{sec:combining}

Given embeddings $\phi(x_n),\phi(x_c)$, we need to combine them so as to obtain a final embedding, denoted $\phi(x)$, which will be used to fit the HD model. There are several reasonable ways to do this, we outline a few below, which are compared empirically in Section \ref{sec:empirical-evaluation}, and found to yield roughly equivalent results.

\vspace{2mm}
\noindent\textbf{Bundling by Concatenation}. The final representation is simply the concatenation of $\phi(x_n)$ and $\phi(x_c)$. This approach may be beneficial because it allows one to easily combine embeddings of different precision and sparsity levels, and because the embeddings may be of different length. This is appropriate, for instance, when there are many more numeric than categorical features (or vice-versa). However, this comes at the expense of a higher final dimension for the encoding, which means that model will have more parameters to estimate.

\vspace{2mm}
\noindent\textbf{Bundling by Sum}. The final representation is the element-wise sum $\phi(x)=\phi(x_n) + \phi(x_c)$. This approach may be desirable because it (1) does not increase the dimension of the final embedding and (2) captures a notion of similarity between the count and categorical embeddings:
\[
    \phi(x)\cdot \phi(x') = \phi(x_n)\cdot\phi(x'_{n})+\phi(x_n)\cdot\phi(x'_{c}) + \cdots + \phi(x_c)\cdot \phi(x'_{c}).
\]
However, a disadvantage of this approach is that the embeddings need to be of the same length, which means that the numeric encodings may need to be of unnecessarily high-dimension to conform with the categorical encodings, or vice-versa. Additionally, the final representation may also require higher precision to store than either of the inputs.

\vspace{2mm}
\noindent\textbf{Bundling by Thresholded-Sum}. The procedure above can be modified by thresholding the sum to ensure the result is of low-precision. For instance, one could threshold $\phi(x) = \phi(x_n) + \phi(x_c)$ at $1$. This method of encoding accords naturally with binary and sparse embeddings, in which case it is equivalent to bundling using the element-wise max, or the logical \texttt{OR}. In this case, assuming the embeddings of the numeric and categorical data to be highly sparse, the probability that both encodings are non-zero on a particular coordinate is small, and so the thresholded sums will be nearly equivalent to the true sum described above, while ensuring the result remains of binary precision.

\section{Hardware Implementation}

In the following sections, we describe implementations of the encoding methods in FPGA and an in-memory architecture.

\subsection{FPGA Implementation}

For our FPGA implementation, we use Xilinx HLS (High-Level Synthesis) which is a C++ based language suited for FPGAs by supporting hardware-related optimization directives \citep{UG1399}.
Our design is flexible in terms of the encoding dimension ($d$), number of numeric and categorical features ($n$ and $s$ respectively), precision of the projection matrix elements ($\Phi$), and degree of hardware parallelism.
Hence, our implementation can be reused for different problems and FPGA sizes by changing the parallelism degree and/or dimension.

\begin{figure}[h]
    \centering
    \includegraphics[width=0.7\textwidth]{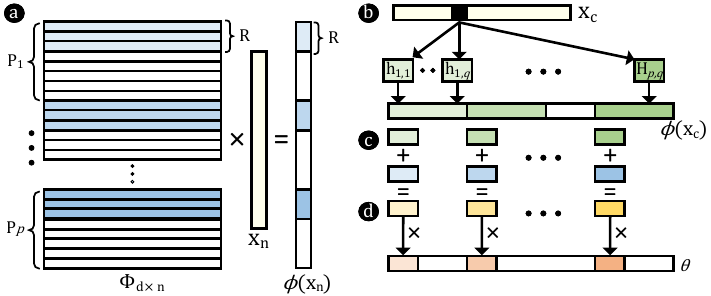}
    \caption{Abstraction of the encoding and update implementation on FPGA. We implemented the design in a dataflow fashion, so the modules operate independently in a producer-consumer manner.}
    \label{fig:flow}
\end{figure}

\vspace{2mm}
\noindent \textbf{Categorical Encoding:} 
To encode a categorical input $x_c$, with $k$ hash-functions $\psi_1$ to $\psi_k$, we need to pass each symbol $a \in x_{c}$ through all of the hash-functions and use the hash output to set (write) the resulting coordinates in the encoding to one.
%Since the dimensionality of the encoding vector $\phi(x_{c})$ is large, we cannot fully partition it.
Since the output of the hash is nondeterministic, the FPGA implementation flow cannot schedule these writes to be parallel.
That is, even if $\phi(x_{c})$ is split into several partitions, we can only have inter-partition parallelisms, whereas it is possible all the $k$ hash outputs point to a certain partition, making all intra-partition writes sequential.
Thus, a basic implementation of sparse hash encoding takes $s \times k \times t_\psi$ cycles, where $t_\psi$ is the effective latency of a hash-function.
To resolve this issue, as shown in Figure \ref{fig:flow}(b), we partition the categorical encoding vector $\phi(x_{c})$ into $p$ partitions, similar to the partitioning of the rows in $\Phi$.
Then, we uniformly split the $k$ hash-functions among these partitions, $q = \sfrac{k}{p}$ hashes per each.
This guarantees each partition does not have more than $q$ write operations, so the hash encoding time reduces to $s \times \sfrac{k}{p} \times t_\psi$ cycles.
We use the Murmur3 hash-function \citep{murmur3} with a $\mathtt{pipeline}$ directive that helps the function to achieve a throughput of one hash per cycle.

\vspace{2mm}
\noindent \textbf{Numeric Encoding:} 
Figure \ref{fig:flow}(a) shows the numeric encoding, which is essentially a $\Phi \times x_n$ matrix-vector multiplication using nested C++ loops over $\Phi$'s rows (outer loop) and columns (inner loop).
We parallelize the numeric encoding by entirely unrolling the loop over the $\Phi$'s columns (i.e., the inner loop).
That is, all $n$ elements of a row in $\Phi$ are multiplied by the numeric elements in $x_{n}$ simultaneously and the dot product result is accumulated in a pipelined manner, making an effective vector-vector multiplication throughput of one cycle.
We partition the $\Phi$ matrix column-wise, so that all elements of a row in $\Phi$ can be read during the same cycle by storing each element in a different on-chip RAM of the FPGA.
In addition, we also partially parallelize over the rows of $\Phi$'s (outer loop), to the extent allowed by the FPGA's resources.
This row-wise unrolling also requires partitioning the $\Phi$'s rows into different block RAMs.
However, we found the capability of the FPGA synthesis flow limited in automated partitioning of the large number of rows in $\Phi$ to a high degree of parallelism.
Thus, according to Figure \ref{fig:flow}(a), we manually partitioned the $\Phi$ into $p$ coarse partitions P$_1$ to P$_p$, and applied a second round of automated partitioning of $R$ rows over each of these partitions.
Thus, effectively, $p \times R$ rows are unrolled.
%Since we could fully unroll each row as alluded above (single cycle vector-vector multiplication), the total latency of $\Phi \times x_n$  becomes $O(\sfrac{d}{5R})$ cycles.
%In our setup using Xilinx Alveo U280 FPGA and $d=10K$ dimensions, we find $R=64{-}128$ optimal in terms of maximizing the resource utilization, yet making the design routable with an acceptable 100--150\,MHz frequency.
Finally, the top-$k$ sparsification of the encoded $\phi(x_{n})$ needs sort operation, which is expensive on FPGA. Thus, following Section \ref{ref:num-sparse}, we instead implement this procedure using thresholding.

\vspace{2mm}
\noindent \textbf{Update (Learning):} 
We pipeline the update step with the encoding step.
For each $p$ partition of $\phi(x_{n})$, $R$ embeddings are generated in one cycle.
As Figure \ref{fig:flow}(c) shows, we combine the numeric and categorical embeddings by sum, threshold-sum, or concatenation and pass to the update module for dot-product with the same elements of the $\theta$ vector.
The $\theta$ vector is also partitioned, and so the dot-product of the encoding and $\theta$ can be done in parallel.

\subsection{Processing-in-Memory Architecture}

\begin{figure}[t]
    \centering
    \includegraphics[width=0.95\textwidth]{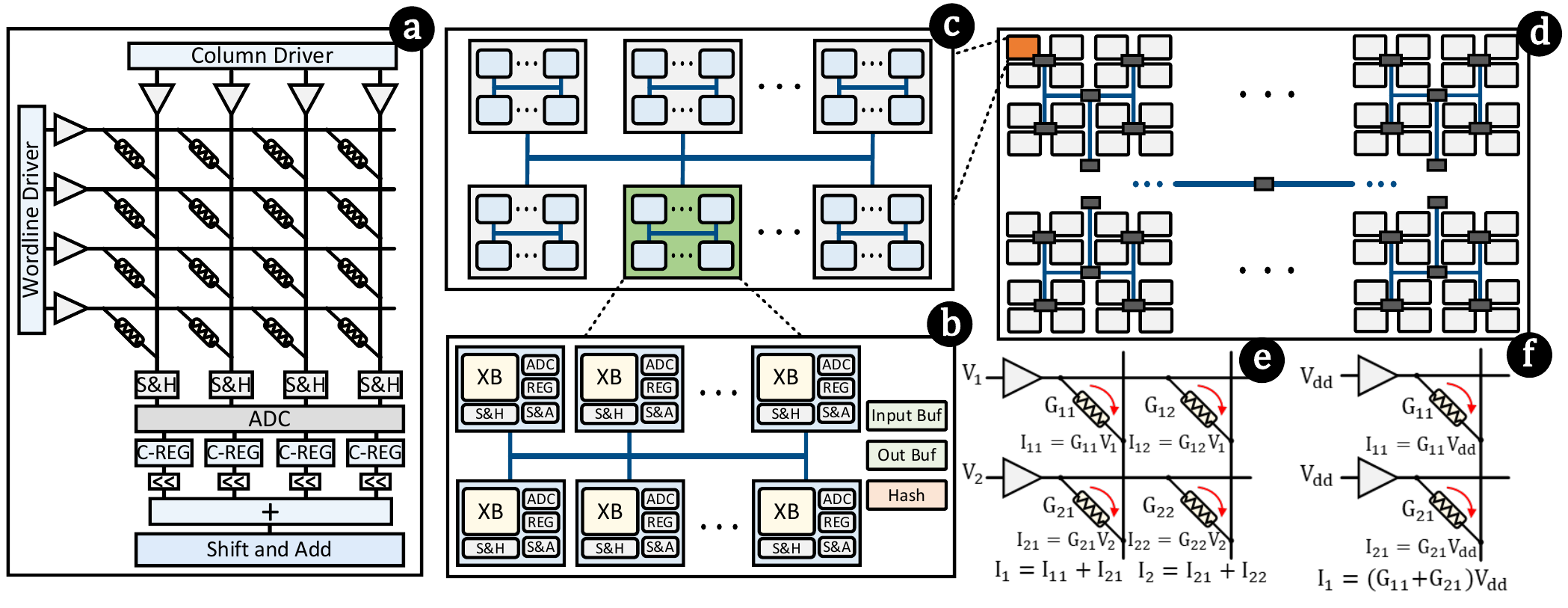}
    \caption{(a) ReRAM storage and compute crossbar, (b)--(d) hierarchical tiled organization of the PIM architecture, and (e)--(f) atomic atomic operations.}
    \label{fig:pim-arch}
\end{figure}

In the following section we describe an implementation of our encoding techniques in an in-memory (PIM) architecture. We first provide a general overview of PIM implementations of primtive operations like element-wise sums and dot-products between vectors, and then explain how these can be specialized to our setting.

\subsubsection{Architecture Overview}
To make the PIM architecture also usable for other applications such as neural networks that can benefit from in-memory computation, we build upon general analog PIM designs based on ReRAM non-volatile memory \textit{crossbar} \citep{shafiee2016isaac}.
The crossbar, shown in Figure \ref{fig:pim-arch}(a), consists of 128 rows and 128 columns (aka bitlines) of cells.
Crossbars with other sizes can also be used with minor changes to the peripheral circuitry.
The principal functionality of the crossbar is aggregating, sensing and digitizing the current that flows through the bitlines.
Essentially, it can count the number of ones at each bitline.
By adding peripheral hardware and decomposing the operations to the bit level, the sensing capability can be used to realize operations such as addition and dot-products between two or more vectors.
The columns of a crossbar are further divided into eight vertical \textit{lanes}, where each lane stores a 16-bit number (128 numbers across all the rows). 
This bit-width is typically sufficient in practice for HDC (embeddings and matrix elements) and other learning-oriented use-cases like CNN weights.

As Figure \ref{fig:pim-arch}(b) shows, a combination of several crossbars forms a \textit{cluster}.
All crossbars of a cluster execute the same instruction (i.e., SIMD processing).
A long vector might span over crossbars of one or more clusters.
Intra-cluster data movements, e.g., writing a crossbar results to output buffer, are routed through a shared bus.
The crossbars contain input and output registers to temporarily store data, and share a decoding and a hashing unit.
Multiple clusters construct a \textit{tile}, which is shown in Figure \ref{fig:pim-arch}(c).
Tiles are connected through a low-cost network such as H-Tree \citep{fujiki2018memory} as global data transfer in PIM applications is expected to be infrequent.

\subsubsection{HDC Operations in Memory}

A crossbar can sense and store the current passing through the columns using the sample-and-hold (S\&H) circuitries.
The analog-to-digital converter (ADC) converts the current to digital domain.
The ADC operates with higher frequency than the ReRAM operations, hence, a single module can be shared between all columns of a crossbar in a time-multiplexed fashion for a reasonable number of columns (e.g., 128 bitlines as in our setting).
For wider crossbars, higher-frequency ADC or multiple instances of an ADC can be used.

By applying a voltage of $V$ over an ReRAM cell of conductance $G$, a current proportional to $V \times G$ passes through the cell. 

Therefore, a logical 0/1 can be achieved by programming $G$ to 0/1.
Summing (bundling) the bits that lie in the same bitline can be done in a single memory cycle by activating their rows.
The result of each bitline is sensed and stored in the column register (C-REG), which holds the data before transferring to other blocks.

\begin{figure}[t]
    \centering
    \includegraphics[width=0.7\textwidth]{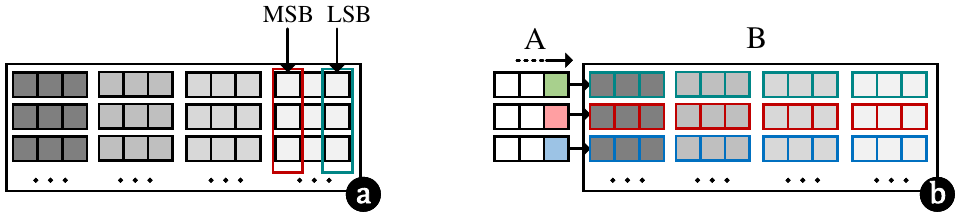}
    \caption{(a) $N$-ary addition, and (b) matrix-vector multiplication. Operands may lay in all or a portion of memory columns.}
    \label{fig:pim-operations}
\end{figure}

\textit{Addition} of $N$ numbers is performed by placing the numbers in $N$ rows of a lane and performing bit-wise accumulation.
Each lane can perform an independent addition as show in Figure \ref{fig:pim-operations}(a).
The result of bit-wise accumulation yields multi-bit partials per bitline, digitized and stored in the column registers of the crossbar.  
%(up to $\log {r_a}{+}1$) 
These partials need another step of summing by passing them through a wired-shift to account for the bit significance of bitlines (i.e., the accumulated result of bit $i$ needs to be shifted left by $i$ indexes), followed by a 16-port tree-adder integrated in each lane.
%The column registers are necessary to temporarily backup the bundling outcome. %before writing back to the memory rows or adding up with the results of the other crossbars in case the vectors do not fit in a crossbar's rows.
Adding all $N$ numbers of a lane is done in two memory cycles; one to activate the $N$ rows and sample the current, followed by digitizing using the time-multiplexed ADC.
Notice that for simple bundling by element-wise sum (as in categorical encoding), each bitline represents a distinct embedding and is independent from the others.
Thus, the mere bundling elides further shifting and accumulating the bitlines results.

\textit{Dot-product} of vectors $A$ and $B$ is implemented by splitting and applying the elements $A$ in a bit-serial fashion, i.e., $ A \cdot B  = \sum_{k}\sum_{i}2^k{A_i[k] B_i}$, where $B$ is placed vertically on the rows of a lane.
As shown in Figure \ref{fig:pim-operations}(b), if $B$ is a matrix, each of its rows is stored on a different lane and the same $A$ can be applied to multiple lanes of a crossbar simultaneously.
Starting from the least-significant bit of $A$, the $i^\text{th}$ element of  vector, $A_i$, is applied to the $i^\text{th}$ element of $B$ to realize $A[0] \cdot B$.
Applying the bits of $A$ happens in the form of 0/1 voltage and acts as a \texttt{AND} operation with the elements of $B$.
The result of applying a certain bit of $A$ is accumulated using the $N$-ary add operation explained above.
In the next cycle, $A[1]$ is applied into $B$, added up and shifted to realize $2^1 \langle A[1] \cdot B \rangle$ and accumulated with the previously stored result.

A dot-product between two $k$-bit vectors or a matrix-vector multiplication takes $k{+}1$ cycles.

\subsubsection{In-Memory Sparse Categorical Data Encoding}

\begin{figure}[t]
    \centering
    \includegraphics[width=0.75\textwidth]{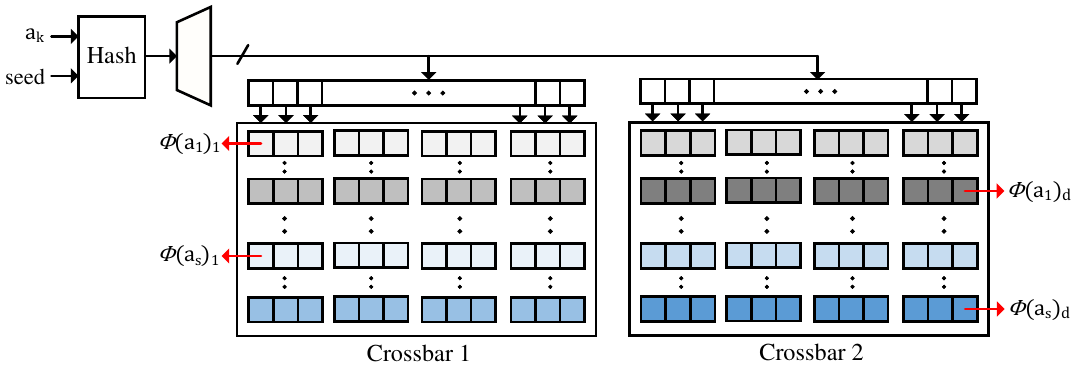}
    \caption{Categorical encoding layout in PIM. In this example, two crossbars are allocated to all $s$ required vectors. Each vector is placed on the allocated crossbars in a row-major fashion. The same index of different vectors lay in the same bitline.}
    \label{fig:enc-categorical}
\end{figure}

Figure \ref{fig:enc-categorical} shows the procedure for sparse hash-encoding of categorical data.
First, the number of crossbars to store $s$ categorical binary vectors are determined and allocated.
Here each \textit{bit} of the crossbar stores one bit of $\phi(a_i)$.
A vector $\phi(a_i)$ has a length of $d$ and spans over all the allocated crossbars in a row-major order, and might be placed in multiple rows of the designated crossbars.
Similar to the FPGA implementation, to facilitate writing sparse bits to the vectors, we logically partition each vector and enforce the criterion that only one bit per partition can be one, the index of which is determined by the hash-function.
The decoder in the Figure \ref{fig:enc-categorical} converts the hash-function output to a one-hot signal to write the proper value into the driving register.
We process the allocated crossbars row by row, starting from the first row of  $\phi(a_1)$.
The hash-function $\psi(a_1)$ determines the single bit index of the row (within all crossbars) that needs to be set to 1.
Thus, processing each physical memory row within all crossbars takes one memory cycle.
By default, we pack all vectors into the minimum number of crossbars.
Thus, all the rows of the designated crossbars are filled, and generating the sparse vector takes $\approx 128$ cycles (one cycle per allocated row).
After that, we activate the proper rows for bundling.
We cannot activate all rows at once since a crossbar stores multiple chunks of the same vector.
In expectation, there are $\sfrac{128}{s}$ different chunks of all vectors in a crossbar that needs separate bundling cycles, so the bundling needs $\sfrac{128}{s}$ cycles.
Note that, depending on the input parameters, we might allocate more than minimal crossbars (e.g., to balance the performance of numeric and categorical encoding) which reduces the number of used rows of crossbars, and therefore, the encoding time.

\subsubsection{In-Memory Numeric Data Encoding}

\begin{figure}[t]
    \centering
    \includegraphics[width=0.75\textwidth]{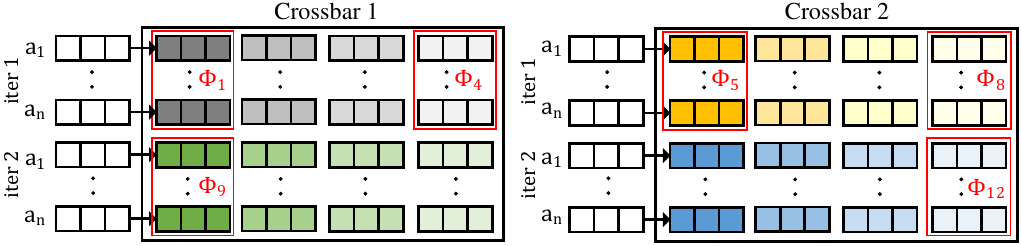}
    \caption{Numeric encoding layout in PIM. In this example, two crossbars are allocated to matrix $\Phi$. The rows of $\Phi$ are placed in the lanes of crossbars in a row-major order.}
    \label{fig:enc-numeric}
\end{figure}

Numeric encoding involves matrix-vector multiplication, which is inherently supported by PIM as explained above.
To realize the numeric encoding $\Phi x_{n}$, we first determine the number of crossbars required to fit $\Phi$. The matrix $\Phi$ then vertically spans the lanes of the allocated crossbars in a row-major order, as illustrated in Figure \ref{fig:enc-numeric}.
Since a row of $\Phi$ may not use up all the rows of a crossbar (as in Figure \ref{fig:enc-numeric}), we can place multiple rows of $\Phi$ within a lane of the crossbar (e.g., $\Phi_1$ and, say, $\Phi_9$ are placed in the same lane).
These rows, however, need separate iterations to be multiplied with the feature vector to avoid the unwanted aggregation of their current in the column registers.
Finally, the outputs of the lanes are transferred to the output register via the shared bus.
%Transferring the outputs of a cluster’s crossbars is done successively via the shared bus.

\section{Empirical Evaluation}
\label{sec:empirical-evaluation}

To understand the practical benefits of hash-based and sparse encodings for HD learning problems, we here apply these methods to a classification task on the ``Criteo'' click-through-rate prediction dataset \citep{criteo}. We emphasize that it is not our goal to achieve state-of-the-art accuracy on this task; we merely wish to understand how hash-encoding compares to conventional techniques in the HDC literature. We selected this dataset simply because it contains, to the best of our knowledge, by far the largest scale of categorical data in any publicly available dataset, and is generally considered to be a challenging classification task, even for state-of-the-art techniques based on deep neural networks \citep{desai2021semantically}.

The data contains information about web advertisements displayed to customers along with a binary label indicating whether or not the advertisement was clicked (e.g. a binary label). The data contains $13$ numeric features, and $26$ categorical features. The features themselves are anonymized, and no information whatsoever is given about their content. The goal is to predict whether or not a customer clicked on an ad. The data comes in two sizes: a ``full'' dataset covering a month of served ads, and a carefully selected (e.g. non-random) subsample that covers one week of ads. The datasets are compared in Table \ref{tab:datasets}. We primarily focus on the smaller $7$-day dataset to keep runtimes tractable when comparing a large number of design choices. We emphasize that the scalability of our approach depends only on the number of features, and the size of the categorical alphabet. Holding these constant, the total number of observations/size of the data is irrelevant from the perspective of computation. Moreover, as shown previously, the encoding dimension in all the compared approaches depends, at worst, only on the logarithm of the alphabet size. Thus, from the perspective of understanding scalablity, there is little difference between the two datasets.

\begin{table}[h]
    \centering
    \begin{tabular}{c|c|c|c}
              & Number of Observations & Size of Categorical Alphabet & Approx. Size on Disk \\ \hline\hline
        Full  & $4.3 \times 10^{9}$ & $1.9 \times 10^{8}$ & 1\,TB\\
        Sampled & $4.6 \times 10^{7}$ & $3.4 \times 10^{7}$ & 10\,GB
    \end{tabular}
    \caption{Comparison of Datasets}
    \label{tab:datasets}
\end{table}

\subsection{Empirical Methodology} \label{subsec:emp-method}

We model the problem using a logistic-regression and estimate parameters using mini-batch stochastic gradient descent (SGD) on the log-likelihood of the data. That is, we estimate $\pr(y = 1) = \sigma(\theta \cdot \phi(x_{n},x_{c}))$, where $\sigma$ is the logistic-sigmoid, $\phi(x_{n},x_{c}) \in \H \subset \R^{d}$ is the HD encoding, and $\theta \in \R^{d}$ is the vector of parameters to be estimated. This approach is similar to the standard HDC ``retraining'' approach based on the perceptron algorithm \citep{imani2017voicehd}, and can also easily be implemented in the on-line setting in which data is streamed continuously.

We use the logistic regression in favor of the perceptron primarily because it comes with precise guarantees on the optimality of the resulting parameters (e.g. $\theta$). The perceptron is merely guaranteed to return \emph{some} linear separator, if one exists. If a separator does not exist, then the result in not guaranteed to be the ``best possible'' separator in any precise sense. Even if one does exist, it may have poor margin (e.g. low robustness to noise in the training data). On the other hand, logistic regression always returns the parameters maximizing the likelihood of the data under a particular statistical model. 

We implement models and fit parameters using TensorFlow. Following standard practice in the literature using this dataset, we use the first $6/7$ of the data (roughly corresponding to $6$ days) for training, and partition the remaining $1/7$ evenly between testing and validation \citep{desai2021semantically}. The validation set is used to tune model parameters like number of hash-functions and encoding dimension and to determine when to stop training the model. Models are validated every $300,000$ records, and we stop training if the loss fails to decrease after $3$ consecutive rounds of validation ($900,000$ observations). Again, following standard practice, we assess model performance using the ``area under the receiver-operating characteristic curve'' (AUC), which better reflects model performance on imbalanced datasets than raw accuracy \citep{desai2021semantically}.

\begin{figure}[h]
    \centering
    \includegraphics[scale=0.4]{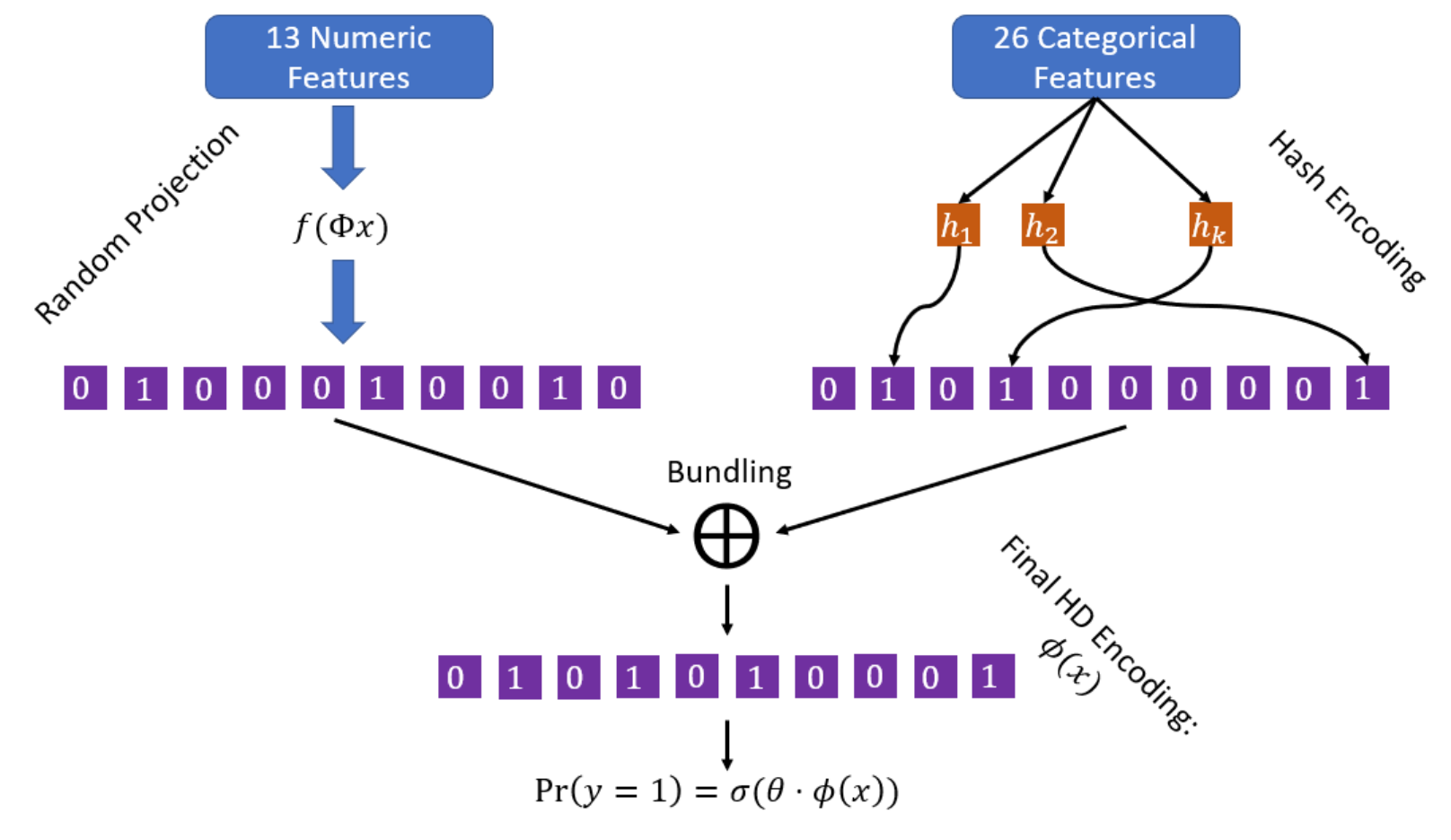}
    \caption{Encoding and model estimation pipeline. Numeric and categorical data are embedded using independent encoding pipelines into a common HD space. The resulting embeddings are combined via a bundling operation and used as input to a classifier.}
    \label{fig:encoding}
\end{figure}

Our encoding and estimation pipeline is shown in Figure \ref{fig:encoding}. The random-projection based encoding methods for numeric data described in section \ref{sec:numeric-encoding}, are implemented as a custom feed-forward layer with no trainable parameters. We implement hash-based encoding methods as a custom Python extension written in \texttt{C++}. We use Murmur3 for the underlying hash-function as it has a well developed \texttt{C} library that is easy to integrate with Python \citep{murmur3,mmh3py}. Each hash-function requires us to store using a random 32-bit integer seed, and so the total space needed to ``store'' the $k$-hash-functions is $32k$ bits. The extension takes as input a mini-batch of categorical samples, and returns the corresponding hash-encoded values which may then be combined with the numeric encodings using any of the methods described in Section \ref{sec:combining}.

\subsection{Results} \label{ref:res}

\subsubsection{Comparing Hash-Based and Random Encoding Methods}

\begin{figure}[h]
    \centering
    \includegraphics[scale=0.4]{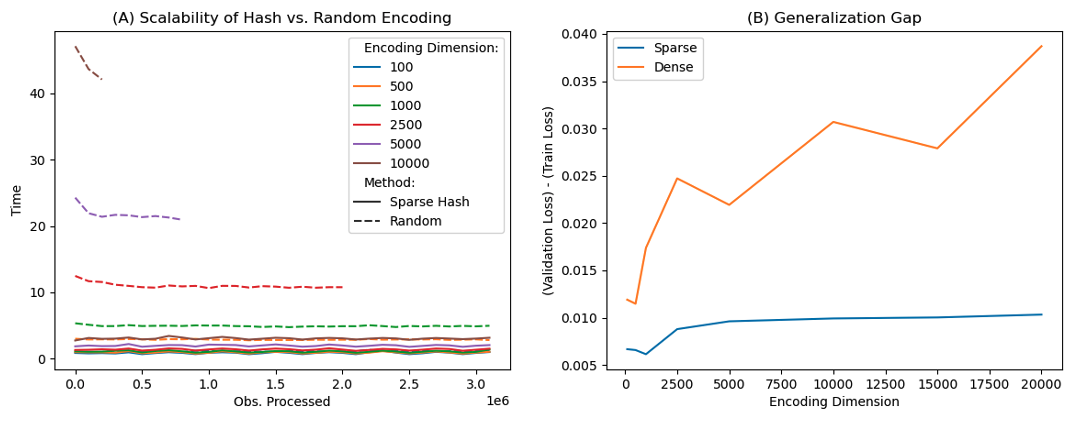}
    \caption{Panel (A) plots the time to encoding batches of $100,000$ observations using naive random encoding generation, and the sparse hashing-based method of Section \ref{sec:hash-encoding}. Panel (B) compares the gap between validation and training loss, for the dense vs. sparse hash based encoding at different values of $d_{\text{cat}}$}
    \label{fig:encoding-time-comparison}
\end{figure}

Figure \ref{fig:encoding-time-comparison} compares the scalability of encoding methods for categorical data. We here measure the time to encode a batch of $100,000$ observations as the encoding dimension is varied. Solid lines indicate the sparse encodings generated using the Bloom filter based method described in Section \ref{sec:hash-encoding}, and dashed lines indicate encodings generated using random-sampling. We do not include the dense hash methods in this plot because they are dramatically slower and would obfuscate the plot. Our random-encoding technique lazily populates a codebook as new symbols are encountered in the data. Thus, the size of the codebook (and the amount of memory required) will increase as more data is processed. To ensure a fair comparison with our \texttt{C} implementation for hash-based encoding, we also implement random-encoding as a custom Python extension written in \texttt{C}, avoiding the high overhead of encoding in native Python.

As can be seen in Figure \ref{fig:encoding-time-comparison}, random encoding generation rapidly encounters scalability bottlenecks as the volume of data processed increases. This is because the categorical alphabet size scales roughly linearly with the number of observations processed, which necessitates storing an ever larger codebook. At a certain point, the codebook size exceeds available RAM, and the program crashes. This problem can be mitigated by using smaller encodings (potentially at the expense of accuracy), or by using caching schemes which retain only the most frequently accessed encodings in memory. However, such approaches do not resolve the fundamental problem, and come with attendant costs in terms of accuracy, latency, and implementation complexity. The naive hash-based method improves on the situation in some ways, since it does not require actually storing the encodings, but it rapidly becomes bottlenecked by computation when $d$ is large. For instance, with $d = 500$, encoding a single batch of data takes about $36$ seconds on a standard CPU machine.

By contrast, in the sparse-hashing based approach, the number of hash-functions remains fixed regardless of the volume of data processed---and the encoding dimension---and so the hash-based encoding methods exhibit constant, high-performance. There is a small overhead from using a larger encoding dimension due to memory allocation, but this is modest. This plot underscores our fundamental observation in this work: encoding techniques that require materializing a codebook simply do not scale to large alphabet sizes. By contrast, hash-based methods offer constant performance independent of the volume of data processed and easily scale to very large data sets.

\subsubsection{Evaluating Hash-Encoding Parameters}

\begin{figure}[h]
    \centering
    \begin{tabular}{c}
        (A) Effect of Number of hash-functions on Model Performance \\
        \includegraphics[scale=0.6]{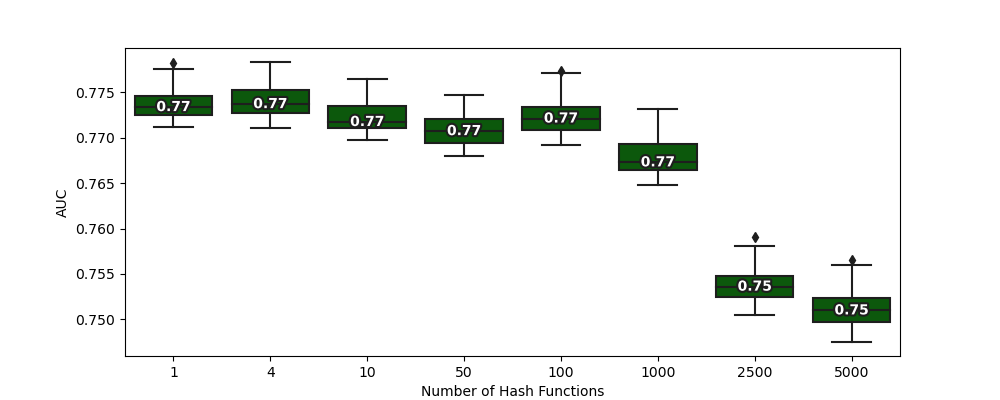} \\
        (B) Effect of Encoding Dimension on Model Performance \\
        \includegraphics[scale=0.6]{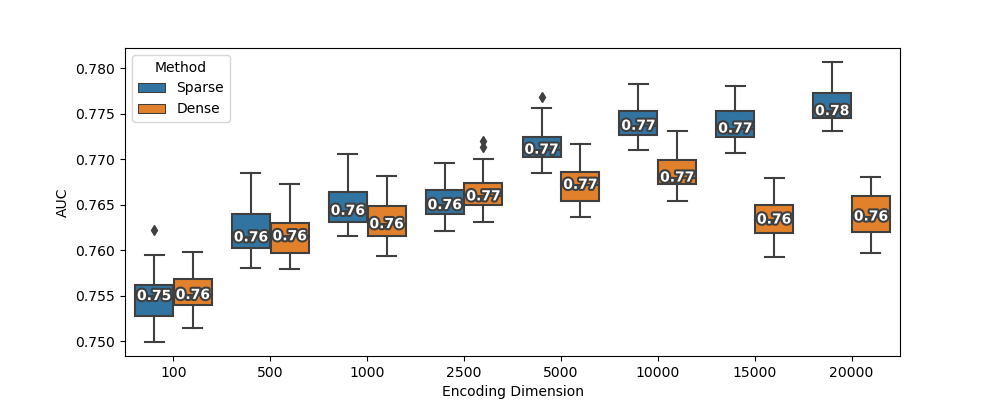}
    \end{tabular}
    \caption{Evaluating the impact of categorical encoding dimension and number of hash-functions on model performance. Box plots show the distribution of AUC on non-overlapping groups of $100,000$ samples. The shaded box indicates the $1$-st and $3$-rd quartile. The solid line indicates the median and the whisker length is $1.5\times$ the IQR. The bundling method is concatenation, the numeric encoding method is a dense random projection ($d=10,000$), and $k = 4$.}
    \label{fig:encoding-dim-comparison}
\end{figure}

We here evaluate the effect of the encoding dimension $d_{\text{cat}}$ for categorical data, and the number of hash-functions/sparsity $k$ on model performance. Results are presented in Figure \ref{fig:encoding-dim-comparison}. For both panels, the numeric encoding method is dense random-projection with fixed $d=10,000$. The bundling method is concatenation, meaning the final model contains $10,000 + d_{\text{cat}}$ parameters. We partition the test and validation sets into chunks, each consisting of $100,000$ samples, and report distributions of model performance, as measured by AUC, as box-plots. The shaded box indicates the $1$-st through $3$-rd quartile, and the solid line indicates the median. The whisker length is $1.5\times$ the inter-quartile range. The number in the box-plot is the median.

Figure \ref{fig:encoding-dim-comparison} (A) compares model AUC as the number of hash-functions is varied, with a constant $d_{\text{cat}} = 10,000$. We find that $k=4$ delivers the best median test error, but that the difference in performance between $k=1$ and $k=100$ is not significant. This is consistent with theoretical results that show error as an increasing function of $k$ for a fixed $d$. This result is appealing from a practical perspective because it means that one can obtain good performance using a handful of hash-functions. Evaluating the hash-functions is the most expensive part of the Bloom filter based encoding method and so reducing $k$ would be expected to lead to better performance. 

To provide context on these results, \citep{desai2021semantically} presents the most recent systematic comparison, to our knowledge, of different deep learning architectures used on this problem and requires $36-540$ million parameters to achieve AUCs of $0.8-0.81$. By contrast, the models in Figure \ref{fig:encoding-dim-comparison} contain $\sim20,000$ (trainable) parameters. We again emphasize that we do not seek to compete with state-of-the-art results on this task. We merely include these comparisons to emphasize that our method falls within the ballpark of results in the literature dedicated specifically to this problem.

Figure \ref{fig:encoding-dim-comparison} (B) presents an analogous plot that fixes $k=4$, and varies the encoding dimension. We here also compare the Bloom filter based encoding method, with the baseline of dense hashing as described in Section \ref{sec:hash-encoding}. Again, consistent with theoretical analysis, performance is strictly increasing for both methods as the dimension is increased. Increasing the categorical encoding dimension results in a consistent increase in AUC up to $d_{\text{cat}} \approx 10,000$ at which point the model saturates and increases in AUC become insignificant. We emphasize that the sparse hash-based encoding methods described here are also advantageous because the number of memory accesses needed for an inference computation depends only on $k$, the number of hash-functions, rather than $d$ the encoding dimension. Accordingly, the increase in accuracy from $d_{\text{cat}} = 500$ to $d_{\text{cat}} = 20,000$ is cheap since the number of hash-functions is held constant.

Interestingly, we find that the sparse hash-encoding method offers markedly better performance for large $d_{\text{cat}}$ than the dense baseline. This is welcome news from a practical perspective since sparse encodings are considerably more efficient computationally, but somewhat surprising given that both methods were shown to preserve dot-products in the sense of Theorem \ref{thm:separability}. We attribute this to a greater propensity for over-fitting for models trained on the dense embeddings, than on the sparse. Figure \ref{fig:encoding-time-comparison} (B) plots the gap between train and validation loss, averaged over the last $10$ rounds of validation, for both the sparse and dense encoding methods. The dense encoding method over-fits with increasing severity as $d_{\text{count}}$ is increased. On the other hand, over-fitting increases very gradually in the sparse representations. This is because only a tiny fraction of the models parameters ($\approx ks/d$) are updated by any given training example--similar in spirit to dropout regularization in deep neural networks. Over-fitting on dense representations could presumably be addressed by L1/L2 regularization, as is in the LASSO model. However, this introduces another hyper-parameter that must be tuned, which is computationally burdensome. The sparse encoding strategy, by contrast, seems to suffer from only very modest over-fitting without the need for any explicit regularization, and are more computationally efficient.

\subsubsection{Comparing Methods for Encoding Numeric Data}

\begin{figure}[h]
    \centering
    \begin{tabular}{c}
    \includegraphics[scale=0.6]{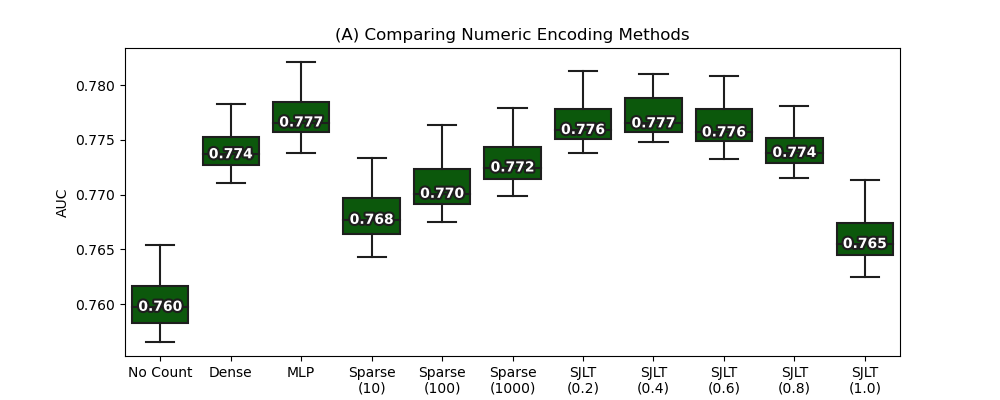}  
    \end{tabular}
    \caption{Figure compares methods for encoding numeric data. ``Dense'' indicates the baseline signed random-projection described in Equation \ref{eqn:dense-rp}. ``Sparse ($k$)'' is the sparse random-projection scheme described in Equation \ref{eqn:sparse-random-projection}, where $k$ is the number of non-zero coordinates in the output.  ``SJLT ($p$)'' indicates the SJLT scheme described in Equation \ref{eqn:sjlt}, where $p$ is the probability that a coordinate in the projection matrix is non-zero. SJLT encodings are quantized using the sign function. ``MLP'' is a simple neural network model, and ``No-Count'' omits numeric data entirely.}
    \label{fig:numeric-comparison}
\end{figure}

Figure \ref{fig:numeric-comparison} (A) compares methods for encoding the numeric data. We compare the dense and sparse random-projection based encoding methods described in Equations \ref{eqn:dense-rp} and \ref{eqn:sparse-random-projection} respectively, along with the SJLT described in Equation \ref{eqn:sjlt}. To simplify implementation of the SJLT, we relax the construction of \citep{kane2014sparser} and simply instantiate $\Phi$ randomly by drawing each coordinate uniformly at random from the distribution:
\[
    \Phi_{ij} = \begin{cases} +1 &\text{ w.p. } p/2 \\ 0 &\text{ w.p. } 1-p \\ -1 &\text{ w.p. } p/2.  \end{cases}
\]
We compare the performance with different choices of $p$ (i.e. number of non-zero components). The categorical encoding method is the sparse ``Bloom filter'' based method, with $d = 10,000$, and $k = 4$. The bundling method for the numeric and categorical encodings is concatenation. The box-and-whisker plots are as described for Figure \ref{fig:encoding-dim-comparison}. 

We compare these approaches against two baselines. The first is to simply omit the numeric data all together, and fit the classifier only on the categorical encodings. The purpose of this baseline is to verify that the count encodings are indeed useful to the classifier. The second is to encode the numeric data using a simple multilayer-perceptron (MLP) style neural network. The MLP contains $4$ hidden layers with $512 \times 256 \times 64 \times 16$ hiden units in each layer for a total of $155,984$ parameters. The MLP is trained along with the logistic-regression classifier using SGD. 

We find that the MLP and SJLT, with a sparsity parameter of $p=0.4$ deliver best results, each achieving a median test AUC of $0.77$ and outperforming the random-projection based methods that rely on dense projection matrices. The SJLT offers two significant advantages: first, it is instantiated randomly at the start of training, and remains fixed from then on. Second, the coordinates in the embedding matrix--$\Phi$--are sparse ($\sim60\%$ zeros), and low-precision. By contrast, the MLP weights are dense and high-precision, and must be trained using back-propagation. A potential disadvantage of the SJLT is that the encodings (e.g. $\phi(x)$) are dense. We find that the sparse random projection method loses just $0.007-0.005$ AUC relative to the SJLT to achieve sparsity levels of $1\%$ and $10\%$ respectively in the encodings, but at the expense of needing to store a dense and high-precision embedding matrix. It would be of interest to study ways to combine these methods. That is, to have both sparse embedding matrices and encodings.

\subsection{Comparing Methods for Bundling Encodings}

\begin{figure}
    \centering
    \includegraphics[scale=0.5]{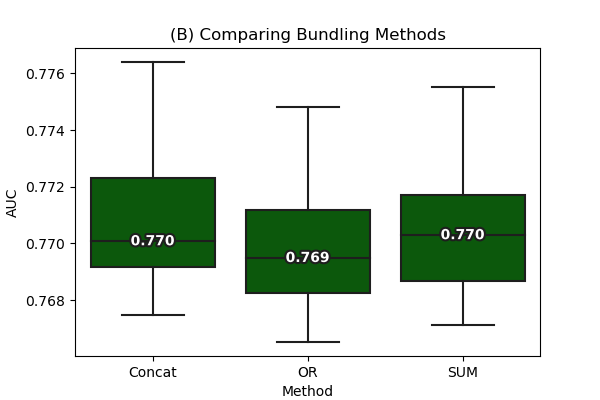}
    \caption{Figure compares methods for bundling the numeric and categorical data as described in Section \ref{sec:combining}. Box plots are as described in Figure \ref{fig:encoding-dim-comparison}. }
    \label{fig:bundle-comparison}
\end{figure}

Figure \ref{fig:bundle-comparison} compares the different methods for combining (or ``bundling'') the encodings of numeric and categorical data described in Section \ref{sec:combining}. The categorical encoding method is the Bloom filter with $d = 10,000$ and $k = 4$, and the numeric encoding method is sparse random projection with $d = 10,000$ and $k = 100$. All three methods yield nearly equivalent performance in terms of AUC. However, bundling by the logical ``or'' is advantageous from a computational perspective since (1) it does not increase the dimension of the bundled representation, and (2) the final embedding is fully binary.

\subsection{Hardware Evaluation}

In the following, we evaluate our FPGA and PIM implementations.

\subsubsection{FPGA Evaluation} \label{ref:eval-fpga}

\noindent \textbf{Setup:}
We implemented the design using Xilinx Vitis HLS 2021.2 \citep{UG1399} on an Alveo U280 Data Center Accelerator Card \citep{u280}.
The FPGA is installed in a machine running Ubuntu 18.04 with Intel Gen-11 Core i7-11700K @4.8\,GHz and 80\,GB of physical memory.
The design supports parameterized vector length $d$, number of numeric ($n$) and categorical ($s$) features, flexible precision of projection matrix, and hardware parallelism level, so that it can be implemented on other FPGA platforms, as well.
We implemented all the three combining techniques, namely thresholded-sum ($\mathtt{OR}$), sum ($\mathtt{SUM}$), and concatenation ($\mathtt{Concat}$), which achieved operating frequencies of 122--150\,MHz.
We also implemented the $\mathtt{No\text{-}Count}$ encoding that omits the numeric data and works at 150\,MHz.
We used five coarse (manual) partition for the projection matrix rows and vectors (i.e., $p=5$ in Figure \ref{fig:flow}), followed by a per-partition parallelism degree of $R = 64$ in the $\mathtt{OR}$ and $\mathtt{SUM}$ combining methods, which makes an effective parallelism of 320.
It means that we can multiply 320 rows of the matrix $\Phi$ with the numeric features per cycle.
The total dimensionality of the $\mathtt{Concat}$ combining is larger ($20K$), so we could set $R = 32$ to avoid routing congestion.
%However, $\mathtt{Concat}$  can work with a relatively higher frequency of 150\,MHz.
The $\mathtt{No\text{-}Count}$ achieved a higher parallelism of $R = 128$ since, without the numeric encoding, this method uses considerably less resources which facilitates more parallelism in partitioning the $\phi(x_c)$ and $\theta$ vectors.

\begin{table}[t]
    \centering
    \resizebox{0.8\textwidth}{!}{
    \begin{tabular}{c|c|c|c|c|c|c}
              & Frequency & $\phi(x_c)$ & $\phi(x_n)$ & $\theta \cdot \phi(x)$ & $\big(y_i - \sigma(\theta \cdot \phi(x)\big) \phi(x)$ & Throughput (M/sec)  \\ \hline\hline
        $\mathtt{OR}$ & 130\,MHz & 31 & 48 & 35 & 34 & 1.51  \\
        $\mathtt{SUM}$ & 122\,MHz &57 & 48 & 40 & 34 & 1.08 \\
        $\mathtt{Concat}$ & 150\,MHz &31 & 80 & 67 & 66 & 0.94 \\
        $\mathtt{No\text{-}Count}$ & 150\,MHz &49 & -- & 20 & 18 & 2.69 \\
    \end{tabular}
  }
        \caption{Frequency, number of cycles (of each step), and throughput (millions of inputs per second) of the FPGA implementation ($d = 10{,}000$).}
    \label{tab:cycle}
\end{table}

\vspace{2mm}
\noindent\textbf{Performance:} 
Table \ref{tab:cycle} reports the cycle count for each of the modules.
Since the encoding and update modules work in a dataflow fashion, the total latency is the maximum of encoding (including both categorical and numeric) and update (calculating the gradients) latency.
The entire design is balanced, so the encoding and update modules take similar latency.
Combining encodings using the $\mathtt{SUM}$ approach takes more cycles than the $\mathtt{OR}$ as the latter only sets a certain subset of coordinates to 1, while in $\mathtt{SUM}$ encoding the embeddings are longer binary in precision.
Thus, an extra read per embedding is needed and the next hashes need to wait for the current result as multiple hash outputs might point to the same index.
The numeric encoding column $\phi(x_n)$ includes the latency of writing to the output FIFO, as well (the $\phi(x_c)$ column in case of $\mathtt{No\text{-}Count}$).
In the $\mathtt{Concat}$ combining technique, both parts of the combined vector work in parallel, but we could set lower ($R=32$) parallelism due to high resource utilization, so the latency of its stages is higher.
On the other hand, $\mathtt{No\text{-}Count}$ is the fastest encoding due to using larger parallelism.
The last column of  Table \ref{tab:cycle} report the throughput of FPGA implementation in terms of million inputs/second (both encoding and learning, i.e., calculating gradient).
The FPGA implementation can process between 939K and 2694K inputs per second. %, while the CPU throughput varies from 5.8K (for $\mathtt{Concat}$) to 18.3K (for $\mathtt{No\text{-}Count}$) inputs per second.
%Specifically, our FPGA implementation gains 155$\times$, 115$\times$, 163$\times$, and 147$\times$ speedup over CPU, respectively, for $\mathtt{OR}$, $\mathtt{SUM}$, $\mathtt{Concat}$, and $\mathtt{No\text{-}Count}$ techniques.
%With the fastest CPU combining method ($\mathtt{OR}$), each epoch of the full Criteo dataset ($4.3 \times 10^9$ inputs) takes over five days, whereas it takes only 47 minutes on our FPGA implementation.
% (seven days with MLP)
Therefore, each epoch of the full Criteo dataset ($4.3 \times 10^9$ inputs) takes only approximately 27 minutes for $\mathtt{No\text{-}Count}$ to 76 minutes for $\mathtt{Concat}$ on FPGA.

\begin{comment}
\begin{figure}[h]
    \centering
    \includegraphics[scale=0.76]{new_hw/throughput.pdf}
    \caption{Performance comparison of the FPGA and CPU implementations ($d = 10{,}000$).}
    \label{fig:throughput}
\end{figure}
\end{comment}

\begin{figure}[t]
    \centering
    \includegraphics[scale=0.66]{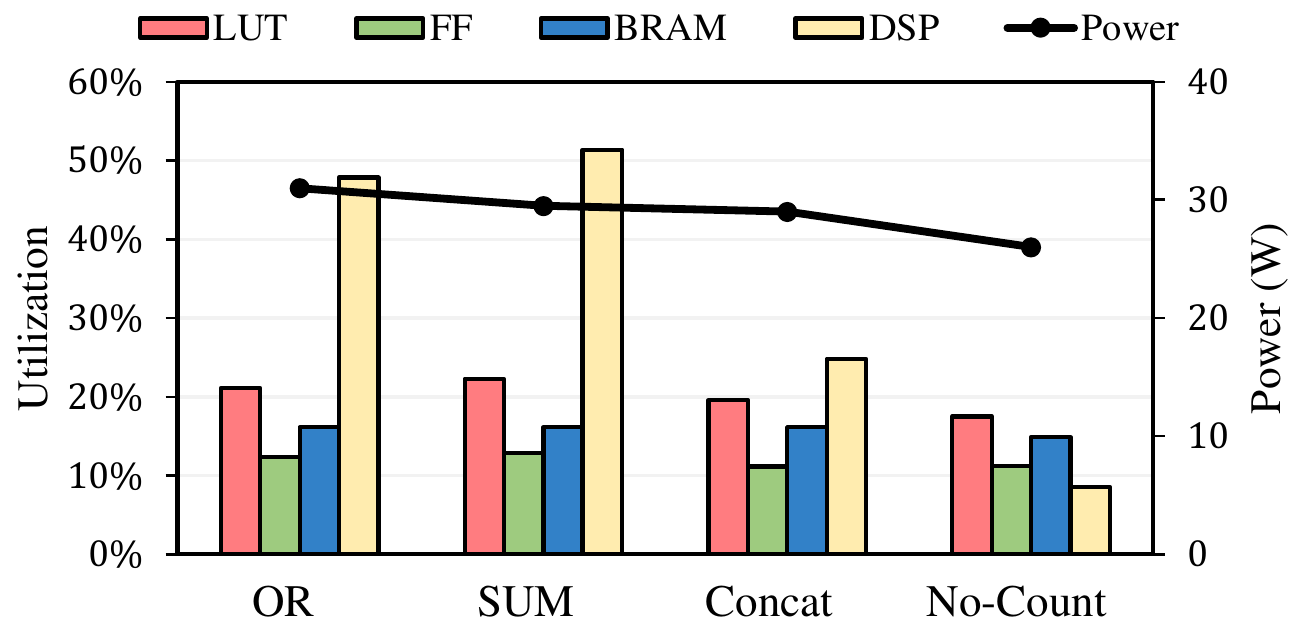}
    \caption{Resource utilization and power consumption for different combining methods ($d = 10{,}000$). The Alveo U280 device contains 1157K Look-Up Tables (LUT), 2384K Flip-Flops (FF), 2016 BRAMs, and 9024 DSPs.}
    \label{fig:resource}
\end{figure}

\vspace{2mm}
\noindent \textbf{Resource and Power Consumption:} 
Figure \ref{fig:resource} shows the FPGA resource utilization for different combining techniques.
$\mathtt{OR}$ and $\mathtt{SUM}$ use a similar amount of resources except $\mathtt{SUM}$ uses slightly more DSPs due to the higher precision of categorical embeddings that require more DSPs when updating the sum.
$\mathtt{Concat}$ uses fewer DSPs due to smaller parallelism ($R=32$), meaning that it performs half of the vector-vector multiplication ($\Phi$ and $x_n$) of the other two.
However, LUT and FF utilization of $\mathtt{Concat}$ is similar to the previous ones as, despite using half parallelism, the total length of the $\mathtt{Concat}$ vectors is twice ($d=20{,}000$), so overall it uses a similar amount of resources.
Finally, $\mathtt{No\text{-}Count}$ does not involve with numeric encoding, so it the least amount of DSPs.

The curve in Figure \ref{fig:resource} (right y-axis) shows the power consumption of the FPGA.
We measured the FPGA power using Alveo's real-time power monitoring. %, and for the CPU we used a Linux power meter tool \citep{CPU-ENERGY-METER}.
The FPGA consumes an idle power of ${\sim}$24\,W which contributes to the major component of the power drain.
Since the resource utilization and operating frequency of the combining approaches is similar, the total power of the FPGA hovers around 30\,W (minimum 26\,W for $\mathtt{No\text{-}Count}$ and maximum 31\,W for $\mathtt{OR}$).
%The CPU power consumption ranges from 85\,W to 90\,W which is ${\sim}3\times$ of the FPGA.
%Thus, the FPGA implementation reduces the energy usage by 422$\times$, 349$\times$, 508$\times$, and 495$\times$ for the $\mathtt{OR}$, $\mathtt{SUM}$, $\mathtt{Concat}$, and $\mathtt{No\text{-}Count}$ approaches, respectively.

\vspace{2mm}
\noindent \textbf{Comparison to Shift-Based Materialization:} Another approach to generate random categorical vectors could be by permuting a set of seed vectors, where a seed vector of length $d$ can create $d$ different vectors \citep{khaleghi2022generic}.
Thus, we can associate each categorical feature vector with a certain permutation of a seed vector, so that the total number of required vectors reduces by a factor of $d$ (e.g., 22.6\,MB for the Criteo dataset).
Selecting the seed vector from the pool of the seeds and then the number of permutations on the selected vector depends on the feature value.
A simple approach to realize it to use two hash-functions over the value of the categorical feature, $\psi_{1}(a)$ and $\psi_{2}(a)$  to determine the seed and the number of permutations.
Since the permutation is variable within $[0, d)$, it requires $O(d)$ cycles to realize ($\sfrac{d}{2}$ on average).
To improve that, we set the permutation granularity to 16 by choosing permutation steps as $\big(\psi_{2}(a) \% \frac{d}{16}\big){\times}16$, meaning that a seed vector can be permuted only in multiples of 16. 
%It increases the number of seed vectors by $16\times$, which is still affordable, but also 
It eases the vector materialization by needing $O(\frac{d}{16})$ cycles to permute, which we further improve it by using data movement instead of permutation.
Once the proper seed vector is read from the FPGA DRAM, we split it into bricks of 16 successive bits.
We initialize a categorical (level) vector, and the hash value $m = \psi_{2}(a) \% \frac{d}{16}$ determines the index from which the \textit{bricks} should be written to the level vector.
Accordingly, the $i^\text{th}$ seed brick goes to the brick number of $(m+i) \% \frac{d}{16}$ of the materialized level vector.
%Since addressed to the level vector are consecutive, we partition the level vector bricks, so that multiple consecutive bricks of this vector can be read and written independently. Thus, the $O(\frac{d}{16})$ latency further reduces by simultaneously writing multiple seed bricks into the level vector.

With all the explained optimizations, we observed that materialization of each level vector (per each categorical feature),  including reading the seed vector from DRAM, takes ${\sim}500$ cycles.
The throughput of this approach, therefore, is limited to ${\sim}$11,200 inputs/second with the categorical encoding being the bottleneck stage in all combining approaches.
Thus, encoding by materialization is $84\times$ slower than our slowest hash-based approach ($\mathtt{Concat}$), and 135$\times$ slower than our hash-based encoding using $\mathtt{OR}$ combining.
%Indeed, even though we accelerated the shift-based materialization on FPGA, its performance is in par with the \textit{CPU} implementation of the sparse hash-based encoding of categorical data. FPGA-accelerated HDC with shift-based materialization is only 14\% (1.14$\times$) faster than the CPU implementation of sparse hash encoding.

\subsubsection{PIM Evaluation} \label{ref:eval-pim}

\begin{table}[t]
    \centering
    \resizebox{0.9\textwidth}{!}{
\begin{tabular}{c|c|c||c|c|c}
Component         & Area ($\mu m^2$) & Power  ($\mu W$) & Component & Area ($\mu m^2$)  & Power ($\mu W$) \\ \hline \hline
128$\times$128 array & 25    & 300   & Hash      & 839    & 8.8   \\ 
ADC               & 570    & 1451    & Decoder   & 26     & 0.02  \\ 
DAC ($\times$256)        & 136   & 5.4  & Router    & 2209   & 459   \\ 
S\&H ($\times$128)       & 5.0   & 1.0   &           &        &       \\ 
Lane peripheral   & 310   & 3.1   & \textbf{Crossbar}  & 3502 $\mu m^2$   & 1.79 mW  \\ 
Output register   & 1646  & 634      & \textbf{Cluster}   & 33042 $\mu m^2$ & 15.9 mW \\ 
Input register    & 2514  & 1011 & \textbf{Tile}      & 0.264 $mm^2$ & 127.6 mW \\ 
Drive  register ($\times$2) & 143   & 2.1     & \textbf{Chip}   & $\mathbf{136\, mm^2}$ & \textbf{65\,W} \\ 
\end{tabular}
}
\caption{PIM components specifications.}
    \label{tab:pim-spec}
\end{table}

\textbf{Setup:} 
Similar to the other PIM designs such as \citep{shafiee2016isaac} and \citep{fujiki2018memory}, we considered $128 \times 128$ crossbars.
Such a crossbar allows sharing an ADC to sense and digitize the current of all 128 bitlines within the 100\,ns read latency of the rows in a time-multiplexed manner.
Table \ref{tab:pim-spec} summarizes the PIM parameters.
For the 128$\times$128 array, sample-and-hold (S\&H), and router, we used the same parameters of \citep{fujiki2018memory}.
We characterized the digital components, i.e., DAC buffers, lane peripheral (including 8-bit column registers, tree-adder, and shift-and-and unit), hash and decoder unit by implementing them in Verilog and synthesizing using 14\,nm standard cell library of GlobalFoundries with Synopsis Design Compiler.
We implemented the Murmur3 hash as a three-stage pipeline.
For the input and output register of the clusters, we used Artisan Memory Compiler using the same process node.
Note that the Alveo U280 FPGA is built on 16\,nm technology, so using a 14\,nm process for the CMOS peripheral is PIM is a fair practice.
For the ADC, we considered the 8-bit ADC fabricated in \citep{kull20133} and scaled the parameters to 14\,nm according to \citep{stillmaker2017scaling} to match the process technology of the other peripherals.
We consider a total PIM capacity of 512\,Mbit, arranged as eight crossbars per cluster, and eight clusters per tile, making total 32,768 crossbars (512 tiles) while keeping the area and power consumption reasonable.
Notice that unlike \citep{shafiee2016isaac}, we consider an 8-bit column register (C-REG) for each bitline.
These registers are necessary to latch the data before adding up all the bitlines' results.
In addition, in case of bundling, each bitline is independent. The column registers are required to hold the data during writing them to the cluster output register using the shared bus.
As a result, area of the \textit{lanes} contribute to 60\% of the total area.
The total CMOS circuity, including the lanes, hash, decoder, registers and routers contribute to 75\% of the area and 12\% of the total power.
The ADCs consume 73\% of the total power.

\vspace{2mm}
\noindent \textbf{Performance:} 
Table \ref{tab:pim-perf} summarizes the performance details of implementing the proposed encoding methods in PIM.
``Allocated crossbars'' column indicates the number of crossbars of the projection ($\Phi$) matrix for numeric encoding, and level vectors for categorical encoding, per input.
``Utilization rate'' shows the percentage of active crossbar rows.
``Encoding cycles'' shows the number of cycles of each encoding approach (each memory cycle takes 100\,ns).
The ``Throughput'' column reports the throughput in terms of million inputs that can be encoded per second using all crossbars of the PIM chip.
The PIM architecture contains 32,768 crossbars, and can simultaneously process large number of inputs, leading to massive throughput.

The $\mathtt{OR}$ and $\mathtt{SUM}$ only differ in quantizing the categorical encoding, which is carried out after the encoding.
Thus, these two encodings exhibit the same crossbar usage and latency.
Notice that the number of allocated crossbars for the categorical encoding in $\mathtt{OR}$ and $\mathtt{SUM}$ encodings is higher than the $\mathtt{No\text{-}Count}$ (40 versus 20) as in the former ones the numeric encoding takes 81 cycles; hence, to keep up with the performance of numeric encoding, more crossbars are designated for the categorical encoding (at the cost of lower utilization rate, i.e., 41\% versus 81\%).
The numeric and categorical encoding are carried out concurrently.
The $\mathtt{No\text{-}Count}$ encoding only performs categorical encoding which needs significantly less resources per input and achieves higher throughput by better input-level parallelism.
Note that the $\mathtt{No\text{-}Count}$ encoding assigns minimum number of crossbars to store the level vectors to improve the utilization rate and maximize the throughput.
In other words, assigning more crossbars decreases the number of cycles, but the overall throughput diminishes.

\begin{table}[h]
    \centering
    \resizebox{0.95\textwidth}{!}{
\begin{tabular}{c|cc|cc|cc|c}
         & \multicolumn{2}{c|}{Allocated Crossbars}    & \multicolumn{2}{c|}{Utilization Rate}           & \multicolumn{2}{c|}{Encoding Cycles}       & \multirow{2}{*}{\begin{tabular}[c]{@{}c@{}}Throughput\\ (M/sec)\end{tabular}} \\ 
         & \multicolumn{1}{c}{Numeric} & Categorical & \multicolumn{1}{c}{Numeric} & Categorical & \multicolumn{1}{c}{Numeric} & Categorical &                    \\ \hline \hline
$\mathtt{OR}$/$\mathtt{SUM}$   & \multicolumn{1}{c|}{144}        &   40          & \multicolumn{1}{c|}{91\%}        &    41\%         & \multicolumn{1}{c|}{81}        &      80	       &     21.97               \\ 
$\mathtt{No\text{-}Count}$ & \multicolumn{1}{c|}{$-$}        &   20          & \multicolumn{1}{c|}{$-$}        &    81\%         & \multicolumn{1}{c|}{$-$}        &         132    &   103.41                 \\ 
\end{tabular}
}
\caption{PIM performance details.}
    \label{tab:pim-perf}
\end{table}

\subsubsection{Comparison with CPU}

\begin{figure}[h]
    \centering
    \includegraphics[scale=0.70]{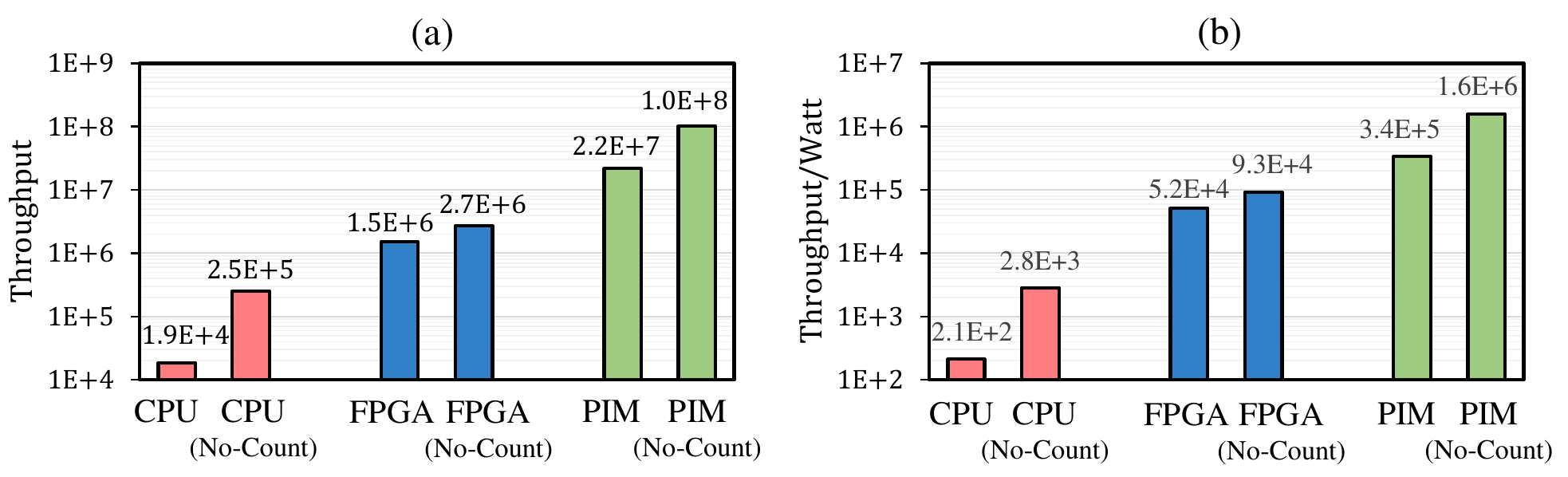}
    \caption{(a) Throughput (inputs per second)  and (b) Throughput/Watt comparison of the encoding step for different platforms. The $\mathtt{No\text{-}Count}$ encoding omits the numeric data.}
    \label{fig:compare_enc}
\end{figure}

\noindent \textbf{Encoding:} 
Figure \ref{fig:compare_enc}(a) shows the encoding throughput in terms of number of inputs encoded in a second.
We performed the CPU experiments using a system with Intel Core i7-8700K 3.70\,GHz CPU with 64\,GB memory.
Notice that performance of encoding step is independent of the subsequent quantization, combining, and learning steps.
The bars labeled as $\mathtt{No\text{-}Count}$ only consider the categorical data.
The PIM results are the same reported in Table \ref{tab:pim-perf}.
When considering both numeric and categorical data, FPGA and PIM achieve 81$\times$ and 1177$\times$ speedup over CPU, respectively.
Without the numeric data (i.e., $\mathtt{No\text{-}Count}$  setting), which makes the CPU encoding relatively faster, FPGA and PIM yield 11$\times$ and 414$\times$ speedup over CPU, respectively.

To account the power consumption differences, Figure \ref{fig:compare_enc}(b) compares the throughput normalized to power (input per second per Watt, representing input per Joule).
From subsections \ref{ref:eval-fpga} and \ref{ref:eval-pim} recap that FPGA and PIM consume a power of 29\,W and 65\,W, respectively.
We estimated the CPU power consumption using CPU energy meter tool \citep{CPU-ENERGY-METER}, and observed an encoding power of 88\,W.
Accordingly, throughput/Watt of FPGA (PIM) over CPU is 246$\times$ (1594$\times$) when considering both categorical and numeric data, and 33$\times$ (560$\times$) in $\mathtt{No\text{-}Count}$ case.

\begin{figure}[h]
    \centering
    \includegraphics[scale=0.70]{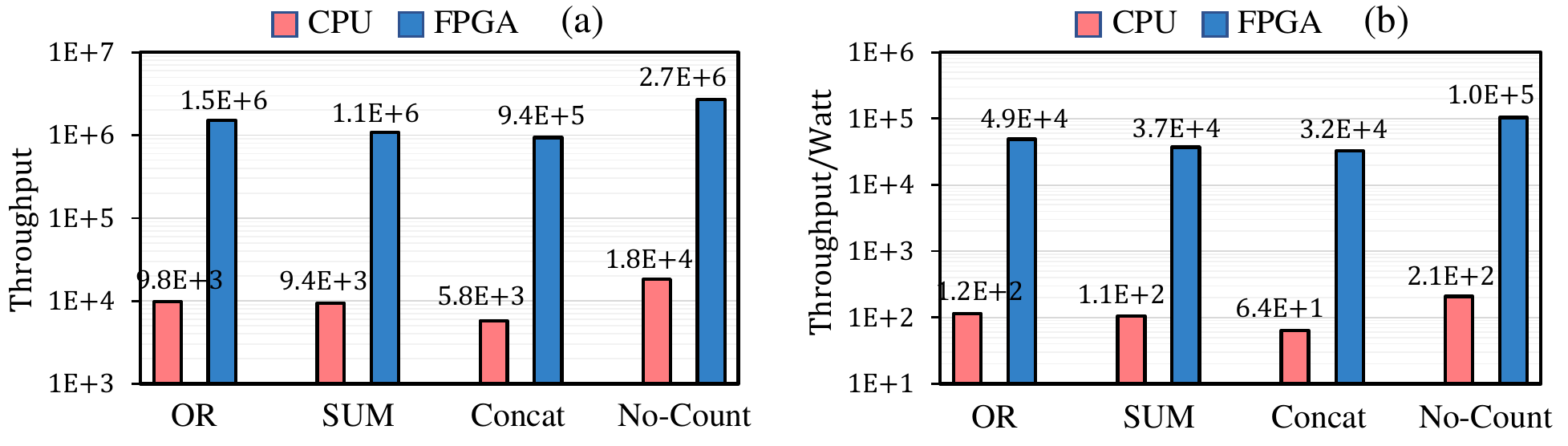}
    \caption{End-to-end (a) throughput (inputs per second)  and (b) throughput/Watt comparison of FPGA and CPU implementation for different combining methods.}
    \label{fig:compare_all}
\end{figure}

\vspace{2mm}
\noindent \textbf{Encoding and Update (Learning):} 
Here we compare the FPGA implementation and CPU for end-to-end learning, i.e., encoding followed by learning through back-propagation.
We do not consider PIM for learning as the learning step entails a lot of data movements and write operations to the memory. We leave it to future work to accelerate learnign based on logistic regression in PIM, although we note that more conventional HD based learning algorithms based on bundling can be effectively implemented in PIM \citep{karunaratne2019memory}. However, these approaches do not provide sufficient statistical power in this context.
Figure \ref{fig:compare_all}(a) compares the end-to-end performance of FPGA and CPU implementations for various combining techniques. The FPGA results are the same as Table \ref{tab:cycle}.
Our FPGA implementation outperforms the CPU by 155$\times$, 115$\times$, 163$\times$, and 147$\times$ for $\mathtt{OR}$, $\mathtt{SUM}$, $\mathtt{Concat}$, and $\mathtt{No\text{-}Count}$ combining techniques, respectively.
In Figure \ref{fig:compare_all}(b), we take the power consumption into account and report the throughput/Watt.
The combining methods almost consume a similar power in each platform.
CPU power hovers around 88\,W. It ranges from 85\,W for $\mathtt{OR}$, up to 90\,W for $\mathtt{Concat}$.
FPGA power consumption ranges from 26\,W in $\mathtt{No\text{-}Count}$ to 31\,W in the $\mathtt{OR}$ technique, making FPGA ${\sim}3\times$ more power-efficient than CPU.
Accordingly, the FPGA implementation delivers 422$\times$, 349$\times$, 508$\times$, 495$\times$ better throughput/Watt than CPU, respectively for $\mathtt{OR}$, $\mathtt{SUM}$, $\mathtt{Concat}$, and $\mathtt{No\text{-}Count}$ combining techniques.
Our FPGA implementation reduces the average energy cost of training one epoch of the Criteo dataset from 1.40 USD to 0.3 cent.

\subsection{Additional Remarks}

To verify that our approach continues to offer reasonable performance even on very large scale data, we run our pipeline on the full 1\,TB version of the data. We use the best performing model architecture from the sub-sampled dataset which uses the SJLT as the numeric encoder, the sparse ``Bloom filter'' categorical encoder, and encoding dimensions of $d_{\text{count}} = 10,000, d_{\text{cat}} = 20,000$. We obtain a median validation AUC of $0.731$ and a test AUC of $0.721$. We attribute the observed difference in performance to additional statistical challenges present in the 1\,TB dataset. The full version of the data is heavily imbalanced--approximately $96\%$ of the records are labelled $0$ (e.g. not-clicked), compared to about $75\%$ in the sub-sampled data. Learning in such heavily skewed settings is challenging in general (even outside of HDC), and we defer addressing this issue to future work since it is orthogonal to our primary concern of scalability in encoding.

\section{Conclusion and Discussion}

In this work we study the problem of obtaining representations of high-dimensional input data in the streaming setting. Our key criteria for these representations is that they (1) be useful for \emph{learning}, and (2) be computationally efficient to compute. Existing encoding methods primarily fail to satisfy the second desideratum because they require maintaining an explicit map between symbols in the alphabet and their HD representations. In high-dimensional settings, existing methods either lead to memory bottlenecks or become burdened by excessive computation. We argue that techniques from the literature on hashing and streaming algorithms from the machine learning literature provide a promising solution to these issues. To the best of our knowledge, we are the first to rigorously develop hash-based encoding methods for use in learning applications based on HDC. 

We present an analytic framework that allows one to compare different methods for encoding high-cardinality data in a learning setting. Our approach naturally relates intrinsic properties of the data, like the geometric separation between classes and the sparsity of the input features, to key parameter choices in the encoding process like the dimension and sparsity of the HD representations. Our work can be seen as a computationally efficient and scalable approach that generalizes existing random encoding strategies via hashing. Indeed, the essence of our approach is to argue that hashing serves as a way to construct the random embeddings ``on-the-fly'' without the need to store any embeddings. To emphasize the difference: encoding a vector consisting of $s$ symbols drawn from an alphabet of size $m$ using standard techniques would require looking up $s$ encodings in a $d \times m$ codebook and then bundling them together to form an HD representation. By contrast, our approach just requires evaluating a small number of hash-functions, depending only logarithmically on the alphabet size, and storing the resulting offsets. Moreover, the resulting embeddings are binary and typically highly sparse, which eliminates multiplications from inference computations.

\section{Acknolwedgements}

This work was supported in part by CRISP, one of six centers in JUMP, an SRC program sponsored by DARPA, and NSF/Intel MLWins grant 2003279.

\clearpage

\vskip 0.2in
\bibliography{refs}

\begin{thebibliography}{77}
\providecommand{\natexlab}[1]{#1}
\providecommand{\url}[1]{\texttt{#1}}
\expandafter\ifx\csname urlstyle\endcsname\relax
  \providecommand{\doi}[1]{doi: #1}\else
  \providecommand{\doi}{doi: \begingroup \urlstyle{rm}\Url}\fi

\bibitem[CPU(2020)]{CPU-ENERGY-METER}
{CPU} energy meter, 2020.
\newblock URL \url{https://github.com/sosy-lab/cpu-energy-meter/}.

\bibitem[u28(2021)]{u280}
Alveo {U280} data center accelerator card data sheet (ds963).
\newblock Datasheet, September 2021.

\bibitem[UG1(2022)]{UG1399}
Vitis high-level synthesis user guide (ug1399).
\newblock User guide, June 2022.

\bibitem[Abadi et~al.(2016)Abadi, Barham, Chen, Chen, Davis, Dean, Devin,
  Ghemawat, Irving, Isard, et~al.]{abadi2016tensorflow}
Mart{\'\i}n Abadi, Paul Barham, Jianmin Chen, Zhifeng Chen, Andy Davis, Jeffrey
  Dean, Matthieu Devin, Sanjay Ghemawat, Geoffrey Irving, Michael Isard, et~al.
\newblock {TensorFlow}: a system for {Large-Scale} machine learning.
\newblock In \emph{12th USENIX symposium on operating systems design and
  implementation (OSDI 16)}, pages 265--283, 2016.

\bibitem[Alonso et~al.(2021)Alonso, Shridhar, Kleyko, Osipov, and
  Liwicki]{alonso2021hyperembed}
Pedro Alonso, Kumar Shridhar, Denis Kleyko, Evgeny Osipov, and Marcus Liwicki.
\newblock Hyperembed: Tradeoffs between resources and performance in {NLP}
  tasks with hyperdimensional computing enabled embedding of n-gram statistics.
\newblock In \emph{2021 International Joint Conference on Neural Networks
  (IJCNN)}, pages 1--9. IEEE, 2021.

\bibitem[Appleby(2016)]{murmur3}
Austin Appleby.
\newblock Murmur3 hash.
\newblock \url{https://github.com/aappleby/smhasher}, 2016.

\bibitem[Babadi and Sompolinsky(2014)]{babadi2014sparseness}
Baktash Babadi and Haim Sompolinsky.
\newblock Sparseness and expansion in sensory representations.
\newblock \emph{Neuron}, 83\penalty0 (5):\penalty0 1213--1226, 2014.

\bibitem[Bloom(1970)]{bloom1970space}
Burton~H Bloom.
\newblock Space/time trade-offs in hash coding with allowable errors.
\newblock \emph{Communications of the ACM}, 13\penalty0 (7):\penalty0 422--426,
  1970.

\bibitem[Boyd and Vandenberghe(2014)]{boyd2004convex}
Stephen~P. Boyd and Lieven Vandenberghe.
\newblock \emph{Convex Optimization}.
\newblock Cambridge University Press, 2014.

\bibitem[Broder and Mitzenmacher(2004)]{broder2004network}
Andrei Broder and Michael Mitzenmacher.
\newblock Network applications of {Bloom} filters: A survey.
\newblock \emph{Internet Mathematics}, 1\penalty0 (4):\penalty0 485--509, 2004.

\bibitem[Broder(1997)]{broder1997resemblance}
Andrei~Z Broder.
\newblock On the resemblance and containment of documents.
\newblock In \emph{Proceedings. Compression and Complexity of SEQUENCES 1997
  (Cat. No. 97TB100171)}, pages 21--29. IEEE, 1997.

\bibitem[Cesa-Bianchi and Gentile(2005)]{bianchi2005improved}
Nicol\`{o} Cesa-Bianchi and Claudio Gentile.
\newblock Improved risk tail bounds for on-line algorithms.
\newblock In \emph{Advances in Neural Information Processing Systems},
  volume~18, 2005.

\bibitem[Chen et~al.(2015)Chen, Wilson, Tyree, Weinberger, and
  Chen]{chen2015compressing}
Wenlin Chen, James Wilson, Stephen Tyree, Kilian Weinberger, and Yixin Chen.
\newblock Compressing neural networks with the hashing trick.
\newblock In \emph{International conference on machine learning}, pages
  2285--2294. PMLR, 2015.

\bibitem[Cohen et~al.(2018)Cohen, Jayram, and Nelson]{cohen2018simple}
Michael~B Cohen, TS~Jayram, and Jelani Nelson.
\newblock Simple analyses of the sparse {Johnson-Lindenstrauss} transform.
\newblock In \emph{1st Symposium on Simplicity in Algorithms (SOSA 2018)}.
  Schloss Dagstuhl-Leibniz-Zentrum fuer Informatik, 2018.

\bibitem[Cormode and Muthukrishnan(2005)]{cormode2005improved}
Graham Cormode and Shan Muthukrishnan.
\newblock An improved data stream summary: the count-min sketch and its
  applications.
\newblock \emph{Journal of Algorithms}, 55\penalty0 (1):\penalty0 58--75, 2005.

\bibitem[{Criteo Research}(2021)]{criteo}
{Criteo Research}.
\newblock Criteo research datasets.
\newblock \url{https://ailab.criteo.com/ressources/}, 2021.
\newblock Accessed: September, 2021.

\bibitem[Dasgupta et~al.(2010)Dasgupta, Kumar, and
  Sarl{\'o}s]{dasgupta2010sparse}
Anirban Dasgupta, Ravi Kumar, and Tam{\'a}s Sarl{\'o}s.
\newblock A sparse {Johnson-Lindenstrauss} transform.
\newblock In \emph{Proceedings of the forty-second ACM symposium on Theory of
  computing}, pages 341--350, 2010.

\bibitem[Dasgupta and Tosh(2020)]{dasgupta2020expressivity}
Sanjoy Dasgupta and Christopher Tosh.
\newblock Expressivity of expand-and-sparsify representations.
\newblock \emph{arXiv preprint arXiv:2006.03741}, 2020.

\bibitem[Dasgupta et~al.(2018)Dasgupta, Sheehan, Stevens, and
  Navlakha]{dasgupta2018neural}
Sanjoy Dasgupta, Timothy~C Sheehan, Charles~F Stevens, and Saket Navlakha.
\newblock A neural data structure for novelty detection.
\newblock \emph{Proceedings of the National Academy of Sciences}, 115\penalty0
  (51):\penalty0 13093--13098, 2018.

\bibitem[Desai et~al.(2021)Desai, Pan, Sun, Chou, and
  Shrivastava]{desai2021semantically}
Aditya Desai, Yanzhou Pan, Kuangyuan Sun, Li~Chou, and Anshumali Shrivastava.
\newblock Semantically constrained memory allocation ({SCMA}) for embedding in
  efficient recommendation systems.
\newblock \emph{arXiv preprint arXiv:2103.06124}, 2021.

\bibitem[Diao et~al.(2021)Diao, Kleyko, Rabaey, and Olshausen]{DiaoGLVQHD2021}
C.~Diao, D.~Kleyko, J.~M. Rabaey, and B.~A. Olshausen.
\newblock {Generalized Learning Vector Quantization for Classification in
  Randomized Neural Networks and Hyperdimensional Computing}.
\newblock In \emph{{International Joint Conference on Neural Networks
  (IJCNN)}}, pages 1--9, 2021.

\bibitem[Frady et~al.(2018)Frady, Kleyko, and Sommer]{frady2018theory}
E~Paxon Frady, Denis Kleyko, and Friedrich~T Sommer.
\newblock A theory of sequence indexing and working memory in recurrent neural
  networks.
\newblock \emph{Neural Computation}, 30\penalty0 (6):\penalty0 1449--1513,
  2018.

\bibitem[Frady et~al.(2021)Frady, Kleyko, and Sommer]{frady2021variable}
Edward~Paxon Frady, Denis Kleyko, and Friedrich~T Sommer.
\newblock Variable binding for sparse distributed representations: theory and
  applications.
\newblock \emph{IEEE Transactions on Neural Networks and Learning Systems},
  2021.

\bibitem[Freedman(1975)]{freedman1975tail}
David~A Freedman.
\newblock On tail probabilities for martingales.
\newblock \emph{the Annals of Probability}, pages 100--118, 1975.

\bibitem[Fujiki et~al.(2018)Fujiki, Mahlke, and Das]{fujiki2018memory}
Daichi Fujiki, Scott Mahlke, and Reetuparna Das.
\newblock In-memory data parallel processor.
\newblock \emph{ACM SIGPLAN Notices}, 53\penalty0 (2):\penalty0 1--14, 2018.

\bibitem[Ge and Parhi(2020)]{ge2020classification}
Lulu Ge and Keshab~K Parhi.
\newblock Classification using hyperdimensional computing: A review.
\newblock \emph{IEEE Circuits and Systems Magazine}, 20\penalty0 (2):\penalty0
  30--47, 2020.

\bibitem[Hastie et~al.(2009)Hastie, Tibshirani, Friedman, and
  Friedman]{hastie2009elements}
Trevor Hastie, Robert Tibshirani, Jerome~H Friedman, and Jerome~H Friedman.
\newblock \emph{The elements of statistical learning: data mining, inference,
  and prediction}, volume~2.
\newblock Springer, 2009.

\bibitem[Imani et~al.(2017{\natexlab{a}})Imani, Kong, Rahimi, and
  Rosing]{imani2017voicehd}
Mohsen Imani, Deqian Kong, Abbas Rahimi, and Tajana Rosing.
\newblock Voicehd: Hyperdimensional computing for efficient speech recognition.
\newblock In \emph{2017 IEEE International Conference on Rebooting Computing
  (ICRC)}, pages 1--8. IEEE, 2017{\natexlab{a}}.

\bibitem[Imani et~al.(2017{\natexlab{b}})Imani, Rahimi, Kong, Rosing, and
  Rabaey]{imani2017exploring}
Mohsen Imani, Abbas Rahimi, Deqian Kong, Tajana Rosing, and Jan~M Rabaey.
\newblock Exploring hyperdimensional associative memory.
\newblock In \emph{2017 IEEE International Symposium on High Performance
  Computer Architecture (HPCA)}, pages 445--456. IEEE, 2017{\natexlab{b}}.

\bibitem[Imani et~al.(2019{\natexlab{a}})Imani, Morris, Messerly, Shu, Deng,
  and Rosing]{imani2019bric}
Mohsen Imani, Justin Morris, John Messerly, Helen Shu, Yaobang Deng, and Tajana
  Rosing.
\newblock {BRIC}: Locality-based encoding for energy-efficient brain-inspired
  hyperdimensional computing.
\newblock In \emph{Proceedings of the 56th Annual Design Automation Conference
  2019}, page~52. ACM, 2019{\natexlab{a}}.

\bibitem[Imani et~al.(2019{\natexlab{b}})Imani, Salamat, Gupta, Huang, and
  Rosing]{imani2019fach}
Mohsen Imani, Sahand Salamat, Saransh Gupta, Jiani Huang, and Tajana Rosing.
\newblock {FACH}: Fpga-based acceleration of hyperdimensional computing by
  reducing computational complexity.
\newblock In \emph{Proceedings of the 24th Asia and South Pacific Design
  Automation Conference}, pages 493--498. ACM, 2019{\natexlab{b}}.

\bibitem[Imani et~al.(2021)Imani, Zou, Bosch, Rao, Salamat, Kumar, Kim, and
  Rosing]{imani2021revisiting}
Mohsen Imani, Zhuowen Zou, Samuel Bosch, Sanjay~Anantha Rao, Sahand Salamat,
  Venkatesh Kumar, Yeseong Kim, and Tajana Rosing.
\newblock Revisiting hyperdimensional learning for {FPGA} and low-power
  architectures.
\newblock In \emph{2021 IEEE International Symposium on High-Performance
  Computer Architecture (HPCA)}, pages 221--234. IEEE, 2021.

\bibitem[Impagliazzo and Zuckerman(1989)]{impagliazzo1989how}
Russell Impagliazzo and David Zuckerman.
\newblock How to recycle random bits.
\newblock In \emph{30th Annual Symposium on Foundations of Computer Science
  (Research Triangle Park, North Carolina)}, pages 248--253. IEEE, 1989.

\bibitem[Impagliazzo et~al.(1989)Impagliazzo, Levin, and
  Luby]{impagliazzo1989pseudo}
Russell Impagliazzo, Leonid~A Levin, and Michael Luby.
\newblock Pseudo-random generation from one-way functions.
\newblock In \emph{Proceedings of the twenty-first annual ACM symposium on
  Theory of computing}, pages 12--24, 1989.

\bibitem[Kane and Nelson(2014)]{kane2014sparser}
Daniel~M Kane and Jelani Nelson.
\newblock Sparser {Johnson-Lindenstrauss} transforms.
\newblock \emph{Journal of the ACM (JACM)}, 61\penalty0 (1):\penalty0 1--23,
  2014.

\bibitem[Kanerva(1988)]{kanerva1988sparse}
Pentti Kanerva.
\newblock \emph{Sparse distributed memory}.
\newblock MIT press, 1988.

\bibitem[Kanerva(2009)]{kanerva2009hyperdimensional}
Pentti Kanerva.
\newblock Hyperdimensional computing: An introduction to computing in
  distributed representation with high-dimensional random vectors.
\newblock \emph{Cognitive Computation}, 1\penalty0 (2):\penalty0 139--159,
  2009.

\bibitem[Karunaratne et~al.(2019)Karunaratne, Gallo, Cherubini, Benini, Rahimi,
  and Sebastian]{karunaratne2019memory}
Geethan Karunaratne, Manuel~Le Gallo, Giovanni Cherubini, Luca Benini, Abbas
  Rahimi, and Abu Sebastian.
\newblock In-memory hyperdimensional computing.
\newblock \emph{arXiv preprint arXiv:1906.01548}, 2019.

\bibitem[Kazemi et~al.(2021)Kazemi, Sharifi, Zou, Niemier, Hu, and
  Imani]{kazemi2021mimhd}
Arman Kazemi, Mohammad~Mehdi Sharifi, Zhuowen Zou, Michael Niemier, X~Sharon
  Hu, and Mohsen Imani.
\newblock {MIMHD}: Accurate and efficient hyperdimensional inference using
  multi-bit in-memory computing.
\newblock In \emph{2021 IEEE/ACM International Symposium on Low Power
  Electronics and Design (ISLPED)}, pages 1--6. IEEE, 2021.

\bibitem[Khaleghi et~al.(2022)Khaleghi, Kang, Xu, Morris, and
  Rosing]{khaleghi2022generic}
Behnam Khaleghi, Jaeyoung Kang, Hanyang Xu, Justin Morris, and Tajana Rosing.
\newblock Generic: highly efficient learning engine on edge using
  hyperdimensional computing.
\newblock In \emph{Proceedings of the 59th ACM/IEEE Design Automation
  Conference}, pages 1117--1122, 2022.

\bibitem[Kim et~al.(2018)Kim, Imani, and Rosing]{kim2018efficient}
Yeseong Kim, Mohsen Imani, and Tajana~Simunic Rosing.
\newblock Efficient human activity recognition using hyperdimensional
  computing.
\newblock In \emph{Proceedings of the 8th International Conference on the
  Internet of Things, {IOT} 2018, Santa Barbara, CA, USA, October 15-18, 2018},
  pages 38:1--38:6. {ACM}, 2018.

\bibitem[Kleyko et~al.(2022)Kleyko, Rachkovskij, Osipov, and
  Rahimi]{KleykoSurveyVSA2021Part2}
D.~Kleyko, D.~A. Rachkovskij, E.~Osipov, and A.~Rahimi.
\newblock {A Survey on Hyperdimensional Computing aka Vector Symbolic
  Architectures, Part {II}: Applications, Cognitive Models, and Challenges}.
\newblock \emph{{ACM Computing Surveys}}, 2022.

\bibitem[Kleyko et~al.(2018)Kleyko, Rahimi, Rachkovskij, Osipov, and
  Rabaey]{kleyko2018classification}
Denis Kleyko, Abbas Rahimi, Dmitri~A Rachkovskij, Evgeny Osipov, and Jan~M
  Rabaey.
\newblock Classification and recall with binary hyperdimensional computing:
  Tradeoffs in choice of density and mapping characteristics.
\newblock \emph{IEEE Transactions on Neural Networks and Learning Systems},
  29\penalty0 (12):\penalty0 5880--5898, 2018.

\bibitem[Kleyko et~al.(2019)Kleyko, Rahimi, Gayler, and
  Osipov]{kleyko2019autoscaling}
Denis Kleyko, Abbas Rahimi, Ross~W Gayler, and Evgeny Osipov.
\newblock Autoscaling {Bloom} filter: controlling trade-off between true and
  false positives.
\newblock \emph{Neural Computing and Applications}, 32:\penalty0 1--10, 2019.

\bibitem[Kleyko et~al.(2020)Kleyko, Rahimi, Gayler, and
  Osipov]{kleyko2020autoscaling}
Denis Kleyko, Abbas Rahimi, Ross~W Gayler, and Evgeny Osipov.
\newblock Autoscaling {Bloom} filter: controlling trade-off between true and
  false positives.
\newblock \emph{Neural Computing and Applications}, 32\penalty0 (8):\penalty0
  3675--3684, 2020.

\bibitem[Kleyko et~al.(2021)Kleyko, Rachkovskij, Osipov, and
  Rahimi]{kleyko2021survey}
Denis Kleyko, Dmitri~A Rachkovskij, Evgeny Osipov, and Abbas Rahimi.
\newblock A survey on hyperdimensional computing aka vector symbolic
  architectures, part {I}: Models and data transformations.
\newblock \emph{ACM Computing Surveys (CSUR)}, 2021.

\bibitem[Kull et~al.(2013)Kull, Toifl, Schmatz, Francese, Menolfi, Braendli,
  Kossel, Morf, Andersen, and Leblebici]{kull20133}
Lukas Kull, Thomas Toifl, Martin Schmatz, Pier~Andrea Francese, Christian
  Menolfi, Matthias Braendli, Marcel Kossel, Thomas Morf, Toke~Meyer Andersen,
  and Yusuf Leblebici.
\newblock A 3.1 mw 8b 1.2 gs/s single-channel asynchronous sar adc with
  alternate comparators for enhanced speed in 32 nm digital soi cmos.
\newblock \emph{IEEE Journal of Solid-State Circuits}, 48\penalty0
  (12):\penalty0 3049--3058, 2013.

\bibitem[Laiho et~al.(2015)Laiho, Poikonen, Kanerva, and
  Lehtonen]{laiho2015high}
Mika Laiho, Jussi~H Poikonen, Pentti Kanerva, and Eero Lehtonen.
\newblock High-dimensional computing with sparse vectors.
\newblock In \emph{2015 IEEE Biomedical Circuits and Systems Conference
  (BioCAS)}, pages 1--4. IEEE, 2015.

\bibitem[Littlestone(1988)]{littlestone1988learning}
Nick Littlestone.
\newblock Learning quickly when irrelevant attributes abound: A new
  linear-threshold algorithm.
\newblock \emph{Machine Learning}, 2\penalty0 (4):\penalty0 285--318, 1988.

\bibitem[Meng et~al.(2016)Meng, Bradley, Yavuz, Sparks, Venkataraman, Liu,
  Freeman, Tsai, Amde, Owen, et~al.]{meng2016mllib}
Xiangrui Meng, Joseph Bradley, Burak Yavuz, Evan Sparks, Shivaram Venkataraman,
  Davies Liu, Jeremy Freeman, DB~Tsai, Manish Amde, Sean Owen, et~al.
\newblock Mllib: Machine learning in {Apache Spark}.
\newblock \emph{The Journal of Machine Learning Research}, 17\penalty0
  (1):\penalty0 1235--1241, 2016.

\bibitem[Mitzenmacher and Vadhan(2008)]{mitzenmacher2008simple}
Michael Mitzenmacher and Salil~P Vadhan.
\newblock Why simple hash functions work: exploiting the entropy in a data
  stream.
\newblock In \emph{SODA}, volume~8, pages 746--755, 2008.

\bibitem[Nisan and Zuckerman(1996)]{nisan1996randomness}
Noam Nisan and David Zuckerman.
\newblock Randomness is linear in space.
\newblock \emph{Journal of Computer and System Sciences}, 52\penalty0
  (1):\penalty0 43--52, 1996.

\bibitem[Plate(2003)]{plate2003holographic}
T.A. Plate.
\newblock \emph{Holographic Reduced Representation: Distributed Representation
  for Cognitive Structures}.
\newblock CSLI Lecture Notes (CSLI- CHUP) Series. CSLI Publications, 2003.
\newblock ISBN 9781575864303.
\newblock URL \url{https://books.google.com/books?id=cKaFQgAACAAJ}.

\bibitem[Rachkovskij(2015{\natexlab{a}})]{rachkovskij2015estimation}
DA~Rachkovskij.
\newblock Estimation of vectors similarity by their randomized binary
  projections.
\newblock \emph{Cybernetics and Systems Analysis}, 51:\penalty0 808--818,
  2015{\natexlab{a}}.

\bibitem[Rachkovskij(2015{\natexlab{b}})]{rachkovskij2015formation}
DA~Rachkovskij.
\newblock Formation of similarity-reflecting binary vectors with random binary
  projections.
\newblock \emph{Cybernetics and Systems Analysis}, 51\penalty0 (2):\penalty0
  313--323, 2015{\natexlab{b}}.

\bibitem[Rachkovskij and Fedoseyeva(1990)]{rachkovskij1990audio}
DA~Rachkovskij and TV~Fedoseyeva.
\newblock On audio signals recognition by multilevel neural network.
\newblock In \emph{Proceedings of The International Symposium on Neural
  Networks and Neural Computing-NEURONET'90}, pages 281--283, 1990.

\bibitem[Rachkovskij et~al.(2012)Rachkovskij, Misuno, and
  Slipchenko]{rachkovskij2012randomized}
DA~Rachkovskij, IS~Misuno, and SV~Slipchenko.
\newblock Randomized projective methods for the construction of binary sparse
  vector representations.
\newblock \emph{Cybernetics and Systems Analysis}, 48:\penalty0 146--156, 2012.

\bibitem[Rachkovskij and Kussul(2001)]{rachkovskij2001binding}
Dmitri~A Rachkovskij and Ernst~M Kussul.
\newblock Binding and normalization of binary sparse distributed
  representations by context-dependent thinning.
\newblock \emph{Neural Computation}, 13\penalty0 (2):\penalty0 411--452, 2001.

\bibitem[Raginsky and Lazebnik(2009)]{raginsky2009locality}
Maxim Raginsky and Svetlana Lazebnik.
\newblock Locality-sensitive binary codes from shift-invariant kernels.
\newblock In \emph{Advances in Neural Information Processing Systems}, pages
  1509--1517, 2009.

\bibitem[Rahimi et~al.(2017{\natexlab{a}})Rahimi, Datta, Kleyko, Frady,
  Olshausen, Kanerva, and Rabaey]{rahimi2017high}
Abbas Rahimi, Sohum Datta, Denis Kleyko, Edward~Paxon Frady, Bruno Olshausen,
  Pentti Kanerva, and Jan~M Rabaey.
\newblock High-dimensional computing as a nanoscalable paradigm.
\newblock \emph{IEEE Transactions on Circuits and Systems I: Regular Papers},
  64\penalty0 (9):\penalty0 2508--2521, 2017{\natexlab{a}}.

\bibitem[Rahimi et~al.(2017{\natexlab{b}})Rahimi, Tchouprina, Kanerva,
  Mill{\'a}n, and Rabaey]{rahimi2017hyperdimensional}
Abbas Rahimi, Artiom Tchouprina, Pentti Kanerva, Jos{\'e} del~R Mill{\'a}n, and
  Jan~M Rabaey.
\newblock Hyperdimensional computing for blind and one-shot classification of
  eeg error-related potentials.
\newblock \emph{Mobile Networks and Applications}, 25:\penalty0 1--12,
  2017{\natexlab{b}}.

\bibitem[Rahimi et~al.(2018)Rahimi, Kanerva, Benini, and
  Rabaey]{rahimi2018efficient}
Abbas Rahimi, Pentti Kanerva, Luca Benini, and Jan~M Rabaey.
\newblock Efficient biosignal processing using hyperdimensional computing:
  Network templates for combined learning and classification of {ExG} signals.
\newblock \emph{Proceedings of the IEEE}, 107\penalty0 (1):\penalty0 123--143,
  2018.

\bibitem[Rahimi and Recht(2008)]{rahimi2008random}
Ali Rahimi and Benjamin Recht.
\newblock Random features for large-scale kernel machines.
\newblock In \emph{Advances in Neural Information Processing systems}, pages
  1177--1184, 2008.

\bibitem[Rosenblatt(1958)]{rosenblatt.58}
F.~Rosenblatt.
\newblock The {Perceptron}: A probabilistic model for information storage and
  organization in the brain.
\newblock \emph{Psychological Review}, 65\penalty0 (6):\penalty0 386--408,
  1958.

\bibitem[Salamat et~al.(2019)Salamat, Imani, Khaleghi, and
  Rosing]{salamat2019f5}
Sahand Salamat, Mohsen Imani, Behnam Khaleghi, and Tajana Rosing.
\newblock {F5-HD}: Fast flexible {FPGA}-based framework for refreshing
  hyperdimensional computing.
\newblock In \emph{Proceedings of the 2019 ACM/SIGDA International Symposium on
  Field-Programmable Gate Arrays}, pages 53--62, 2019.

\bibitem[Schmuck et~al.(2019)Schmuck, Benini, and Rahimi]{manuel2019hardware}
Manuel Schmuck, Luca Benini, and Abbas Rahimi.
\newblock Hardware optimizations of dense binary hyperdimensional computing:
  Rematerialization of hypervectors, binarized bundling, and combinational
  associative memory.
\newblock \emph{ACM Journal on Emerging Technologies in Computing Systems
  (JETC)}, 15\penalty0 (4):\penalty0 1--25, 2019.

\bibitem[Senuma(2021)]{mmh3py}
Hajime Senuma.
\newblock mmh3 python package.
\newblock \url{https://pypi.org/project/mmh3/}, 2021.

\bibitem[Shafiee et~al.(2016)Shafiee, Nag, Muralimanohar, Balasubramonian,
  Strachan, Hu, Williams, and Srikumar]{shafiee2016isaac}
Ali Shafiee, Anirban Nag, Naveen Muralimanohar, Rajeev Balasubramonian,
  John~Paul Strachan, Miao Hu, R~Stanley Williams, and Vivek Srikumar.
\newblock Isaac: A convolutional neural network accelerator with in-situ analog
  arithmetic in crossbars.
\newblock \emph{ACM SIGARCH Computer Architecture News}, 44\penalty0
  (3):\penalty0 14--26, 2016.

\bibitem[Shaltiel(2011)]{shaltiel2011introduction}
Ronen Shaltiel.
\newblock An introduction to randomness extractors.
\newblock In \emph{International colloquium on automata, languages, and
  programming}, pages 21--41. Springer, 2011.

\bibitem[Shi et~al.(2009)Shi, Petterson, Dror, Langford, Smola, and
  Vishwanathan]{shi2009hash}
Qinfeng Shi, James Petterson, Gideon Dror, John Langford, Alex Smola, and SVN
  Vishwanathan.
\newblock Hash kernels for structured data.
\newblock \emph{Journal of Machine Learning Research}, 10\penalty0 (11), 2009.

\bibitem[Stettler and Axel(2009)]{stettler2009representations}
Dan~D Stettler and Richard Axel.
\newblock Representations of odor in the piriform cortex.
\newblock \emph{Neuron}, 63\penalty0 (6):\penalty0 854--864, 2009.

\bibitem[Stillmaker and Baas(2017)]{stillmaker2017scaling}
Aaron Stillmaker and Bevan Baas.
\newblock Scaling equations for the accurate prediction of cmos device
  performance from 180 nm to 7 nm.
\newblock \emph{Integration}, 58:\penalty0 74--81, 2017.

\bibitem[Thomas et~al.(2021)Thomas, Dasgupta, and
  Rosing]{thomas2020theoretical}
Anthony Thomas, Sanjoy Dasgupta, and Tajana Rosing.
\newblock A theoretical perspective on hyperdimensional computing.
\newblock \emph{Journal of Artificial Intelligence Research}, 72:\penalty0
  215--249, 2021.

\bibitem[Vadhan et~al.(2012)]{vadhan2012pseudorandomness}
Salil~P Vadhan et~al.
\newblock Pseudorandomness.
\newblock \emph{Foundations and Trends{\textregistered} in Theoretical Computer
  Science}, 7\penalty0 (1--3):\penalty0 1--336, 2012.

\bibitem[Vapnik(1998)]{vapnik1998support}
Vladimir Vapnik.
\newblock The support vector method of function estimation.
\newblock In \emph{Nonlinear Modeling}, pages 55--85. Springer, 1998.

\bibitem[Vershynin(2018)]{vershynin2018high}
Roman Vershynin.
\newblock \emph{High-dimensional probability: An introduction with applications
  in data science}, volume~47.
\newblock Cambridge university press, 2018.

\bibitem[Wu et~al.(2018)Wu, Li, Huang, Rahimi, Hills, Hodson, Hwang, Rabaey,
  Wong, Shulaker, et~al.]{wu2018hyperdimensional}
Tony~F Wu, Haitong Li, Ping-Chen Huang, Abbas Rahimi, Gage Hills, Bryce Hodson,
  William Hwang, Jan~M Rabaey, H-S~Philip Wong, Max~M Shulaker, et~al.
\newblock Hyperdimensional computing exploiting carbon nanotube fets, resistive
  ram, and their monolithic 3d integration.
\newblock \emph{IEEE Journal of Solid-State Circuits}, 53\penalty0
  (11):\penalty0 3183--3196, 2018.

\end{thebibliography}

\section{Appendix}

\subsection{Proof of Theorem \ref{thm:separability}}
\label{app:proof-thm-separability}

\begin{proof}
Let $p \in \texttt{conv}(\Z)$ and $q \in \texttt{conv}(\Z')$ be the closest pair of points on the convex hulls of $\Z$ and $\Z'$. By Caratheodry's theorem, there exists a set of $n \leq m+1$ points $\{x_{1},...,x_{n}\} \subset \Z$, and weights $\{\alpha_{1},...,\alpha_{n}\} \subset \R^{n}$, such that:
\[
    p = \sum_{i=1}^{n} \alpha_{i}x_{i}, \text{ and } \sum_{i=1}^{n} \alpha_{i} = 1.
\]
Likewise, there exists a set of $n' \leq m+1$ points, $\{x_{1}',...,x_{n'}'\} \subset \Z'$ and $\{\beta_{1},...,\beta_{n}\} \subset \R^{n}$, such that:
\[
    q = \sum_{i=1}^{n'} \beta_{i}x_{i}', \text{ and } \sum_{i=1}^{n'} \beta_{i} = 1.
\]
Now, let us define:
\[
    \phi(p) = \sum_{i=1}^{n}\alpha_{i}\phi(x_{i}), \text{ and } \phi(q) = \sum_{i=1}^{n'}\beta_{i}\phi(x_{i}').
\]
Applying definition \ref{defn:distance-preservation}, we first observe that:
\begin{align*}
    \|\phi(p)\|_{2}^{2} &= \phi(p)\cdot\phi(p) = \sum_{ij}\alpha_{i}\alpha_{j}\left(\phi(x_{i})\cdot\phi(x_{j}')\right) \\
    &\geq \sum_{ij}\alpha_{i}\alpha_{j}\left(x_{i}\cdot x_{j} - \Delta(d)\right) \\
    &= \|p\|_{2}^{2} - \Delta(d).
\end{align*}
Analogously, $\|\phi(q)\|_{2}^{2} \leq \|q\|_{2}^{2} + \Delta(d)$. Let us define:
\[
    \theta = \phi(p) - \phi(q), \text{ and } \nu = -\frac{1}{2}\left(\|\phi(p)\|_{2}^{2} - \|\phi(q)\|_{2}^{2}\right).
\]
Then, fixing some arbitrary $x_{o} \in \Z$, we can see:
\begin{align*}
    \phi(x_{o}) \cdot \theta + \nu &= \sum_{i=1}^{n} \alpha_{i}\left(\phi(x_{o})\cdot \phi(x_{i})\right) - \sum_{i=1}^{n'} \beta_{i}\left(\phi(x_{o}) \cdot \phi(x_{i}')\right) - \frac{1}{2}\left(\|\phi(p)\|_{2}^{2} - \|\phi(q)\|_{2}^{2}\right) \\
    &\geq \sum_{i}\alpha_{i}(x_{i}\cdot x_{o} - \Delta(d)) - \sum_{i}\beta_{i}(x_{i}' \cdot x_{o} + \Delta(d)) - \frac{1}{2}\left(\|p\|_{2}^{2} - \|q\|_{2}^{2} + 2\Delta(d)\right) \\
    &= x_{o} \cdot (p - q) - \frac{1}{2}(\|p\|_{2}^{2} - \|q\|_{2}^{2}) - 3\Delta(d).
\end{align*}
By a standard proof of the Hyperplane Separation Theorem (e.g. \citep{boyd2004convex}):
\[
    x_{o} \cdot (p - q) - \frac{1}{2}(\|p\|_{2}^{2} - \|q\|_{2}^{2}) \geq \frac{1}{2}\|p - q\|_{2}^{2}, \text{ for all } x_{o} \in \texttt{conv}(\Z).
\]
Therefore, we conclude:
\[
    \phi(x_{o}) \cdot \theta + \nu > 0, \text{ for all } x_{o} \in \texttt{conv}(\Z),
\]
provided:
\[
    \Delta(d) \leq \frac{\gamma}{6}.
\]
The proof for $x_{o}' \in \Z'$ is analogous.
\end{proof}

\subsection{Proof of Theorem \ref{thm:dense-codes}}
\label{app:proof-dense-codes}

The proof uses the following Bernstein inequality which may be found in \citep[Theorem 2.8.4]{vershynin2018high}:
\begin{theorem} \textbf{(Bernstein's Inequality)}
    Let $X_{1},...,X_{n}$ be a collection of independent mean-zero random variables such that $|X_{i}| \leq K$ for all $i$. Then, for every $t > 0$:
    \[
        \pr\left(\left|\sum_{i=1}^{n} X_{i}\right| \geq t\right) \leq 2\exp\left(-\frac{t^{2}/2}{\sigma^{2} + Kt/3}\right),
    \]
    where $\sigma^{2} = \sum_{i=1}^{n}\var(X_{i})$.
\end{theorem}
We may now prove the main result:
\begin{proof} \textbf{of Theorem \ref{thm:dense-codes}} 
Fix some pair $x_{c},x_{c}' \subset \A$, let $\phi(x)_{i}$ denote the $i$-th coordinate of $\phi(x)$, and let $\mathcal{I} = x_{c}\cap x_{c}'$. Then, for any $i \in [d]$:
\begin{align*}
    \phi(x_{c})_{i}\phi(x_{c}')_{i} &= \sum_{a \in \I} \phi(a)_{i}^{2} + 2\sum_{a,a' \in \binom{\I}{2}} \phi(a)_{i}\phi(a')_{i} + \sum_{\substack{a \in \I \\ a' \in (x_{c}\cup x_{c}')\setminus \I}} \phi(a)_{i}\phi(a')_{i} + \sum_{\substack{a \in x_{c}\setminus\I \\ x_{c}' \in x_{c}' \setminus \I }} \phi(a)_{i}\phi(a')_{i} \\
    &= |x_{c} \cap x_{c}'| + Z_{i},
\end{align*}
where $\binom{\I}{2}$ denotes the set of all distinct pairs of symbols in $\I$. Noting that, for any distinct pair $a \neq a' \in \A$, $\E[\phi(a)_{i}\phi(a)_{i}'] = 0$, we conclude:
\[
    \E[\phi(x_{c})_{i}\phi(x_{c}')_{i}] = |x_{c}\cap x_{c}'|.
\]
Now it remains to show concentration around this value. Let us consider the centered random variable $Z_{i} = \phi(x_{c})_{i}\phi(x_{c}')_{i} - |x_{c}\cap x_{c}'|$. Since the terms in $Z_{i}$ are at least pairwise independent, we may decompose the variance over the sum as:
\begin{align*}
    \var(Z_{i}) = 4\sum_{a,a' \in \binom{\I}{2}} \var(\phi(a)_{i}\phi(a')_{i}) + \sum_{\substack{a \in \I \\ a' \in (x_{c}\cup x_{c}')\setminus \I}} \var(\phi(a)_{i}\phi(a')_{i}) + \sum_{\substack{a \in x_{c}\setminus\I \\ x_{c}' \in x_{c}' \setminus \I }} \var(\phi(a)_{i}\phi(a')_{i}) \leq 4s^{2},
\end{align*}
since, for any distinct $a,a'$, $\var(\phi(a)_{i}\phi(a')_{i}) = 1$, and there are at most $s^{2}$ terms in the sum. Moreover, $|Z_{i}| \leq 2s^{2}$, and so, by a Bernstein inequality, for any $t > 0$:
\begin{align*}
    \pr\left(\left|\sum_{i=1}^{d} Z_{i} \right| \geq dt \right) &\leq 2\exp\left(-\frac{d^{2}t^{2}/2}{4ds^{2} + 2s^{2}dt/3}\right) \\
    &= 2\exp\left(- \frac{dt^{2}/2}{2s^{2}(2 + t/3)}\right).
\end{align*}
To guarantee this quantity is at most $\epsilon > 0$, it is sufficient to take:
\[
    t \geq \max\left\{\sqrt{\frac{16s^{2}}{d}\log\frac{2}{\epsilon}}, \frac{8s^{2}}{3d}\log \frac{2}{\epsilon}\right\}.
\]
The result follows by union-bounding over all $\binom{\binom{m}{s}}{2} < m^{2s}/2$ pairs of sets of size $s$, whereupon we may set $\epsilon = 2\delta / m^{2s}$, in which case $\log(2/\epsilon) \leq 2s\log(m/\delta)$. Since either case in the max above implies we must restrict attention to the regime in which $d = \Omega((s^{3}\log m)/d)$, we conclude that, with probability at least $1-\delta$:
\[
    \frac{1}{d}(\phi(x_{c})\cdot\phi(x_{c}')) - |x_{c} \cap x_{c}'| \leq 4\sqrt{\frac{2s^{3}}{d}\log\frac{m}{\delta}},
\]
as claimed. The lower bound is analogous.
\end{proof}

\subsection{Proof of Theorem \ref{thm:bloom-separability}}
\label{app:proof-bloom-separability}
To prove the Theorem, we will require the following Lemma:
\begin{lemma}
\label{lemma:collisions}
Let $\A = \{a_{1},...,a_{m}\}$ be an alphabet of size $m$, let $\psi_{1},...,\psi_{k}$ be a set of $k$ hash-functions $\A \to [d]$ drawn uniformly at random from an $s$-independent family, and let $\X \subset \A$ be any set of $s < m$-symbols drawn from $\A$. Let $Z$ be the number of distinct values in the $k$ hashes of $s$ symbols. Then, with probability at least $1-\delta$:
\[
    Z \geq sk - \frac{s^{2}k^{2}}{2d} - \max\left\{ \sqrt{\frac{2s^{3}k^{2}}{d}\log\frac{m}{\delta}}, \frac{4s}{3}\log\frac{m}{\delta} \right\},
\]
for all sets $\X \subset \A$ of size $s$.
\end{lemma}
The proof of the Lemma makes use of the following martingale form of Bernstein's inequality, which is due to \citep{freedman1975tail,bianchi2005improved}:
\begin{theorem}
    Let $L_{1},L_{2},...$ be a sequence of random variables, $0 \leq L_{i} \leq 1$. Define the bounded martingale difference sequence $V_{i} = \E[L_{i}\,|\,L_{1},...,L_{i-1}] - L_{i}$, and the associated martingale $S_{n} = \sum_{i=1}^{n} V_{i}$, with conditional variance $K_{n} = \sum_{i=1}^{n} \var(L_{i}|L_{1},...,L_{i-1})$. Then, for all $t > 0$:
    \[
        \pr\left( S_{n} \geq t \right) \leq \exp\left(-\frac{t^{2}/2}{K_{n} + 2t/3}\right).
    \]
\end{theorem}
We now prove Lemma \ref{lemma:collisions}: \\
\begin{proof}
Fix some $\X \subset \A$ of size $s$, and let $X_{1},...,X_{sk}$ denote the outcomes of hashing the symbols in $\X$ in some arbitrary order. Now let us define the random variable:
\[
    Z_{i} = \begin{cases} 1 &\text{ if } X_{i} \neq X_{1}, X_{i} \neq X_{2},...,X_{i} \neq X_{i-1} \\ 0 &\text{ otherwise.} \end{cases}
\]
That is, $Z_{i}$ is $+1$ if $X_{i}$ is distinct from all previous values, and $0$ otherwise. Then, the number of unique values amongst $X_{1},...,X_{sk}$ is given by:
\[
    Z = \sum_{i=1}^{sk} Z_{i}.
\]
Now let $\F_{i} = \sigma(Z_{1},...,Z_{i})$ be the filtration generated by $Z_{1},...,Z_{i}$, and note that:
\[
    \E[Z_{i} | \F_{i-1}] \geq 1 - \frac{i-1}{d},
\]
since probability that $X_{i}$ is distinct from all previous values is lower-bounded in the case that all $X_{1},..,X_{i-1}$ were distinct. Moreover, since $Z_{i}$ is Bernoulli:
\[
    \var(Z_{i} | \F_{i-1}) \leq \frac{i-1}{d}.
\]
And let us define:
\[
    K = \sum_{i=1}^{sk} \var(Z_{i} | \F_{i-1}) \leq \frac{sk(sk - 1)}{2d} \leq \frac{s^{2}k^{2}}{2d}.
\]
Now, form the martingale difference sequence $V_{i} = \E[Z_{i}|\F_{i-1}] - Z_{i}$. Then, we may apply the aforementioned martingale Bernstein inequality to conclude that, for any $t > 0$:
\begin{align*}
\pr\left( \sum_{i} V_{i} \geq t \right) =
    \pr\left(Z \leq \sum_{i} \E[Z_{i}|\F_{i-1}] - t \right) &\leq \exp\left(-\frac{t^{2}}{2K + 2t/3}\right) \\
                                       &\leq \exp\left(-\frac{t^{2}}{\frac{2s^{2}k^{2}}{d} + 2t/3}\right).
\end{align*}
To guarantee this quantity is at most $\epsilon$, it is sufficient to take:
\[
    t \geq \max\left\{ \sqrt{\frac{2s^{2}k^{2}}{d}\log(1/\epsilon)}, \frac{4}{3}\log(1/\epsilon), \right\}
\]
from which the result follows by a union bound over all ${m \choose s}$ possible sets of size $s$, whereupon $\log(1/\epsilon) \leq s\log(m /\delta)$, and observing that:
\[
    \sum_{i} \E[Z_{i} | \F_{i-1}] \geq sk - \frac{s^{2}k^{2}}{2d}.
\]
\end{proof}
The main Theorem follows as a corollary of the previous result: \\
\begin{proof}
\textbf{of Theorem \ref{thm:bloom-separability}}: Fix some $x_{c},x_{c}'$, and, as in Lemma \ref{lemma:collisions}, let $Z$ denote the number of unique values among the hashes of the symbols in $|x_{c} \cup x_{c}'|$. Suppose that $Z = |x_{c} \cup x_{c}'|$ -- that is, the hashes of all symbols in $x_{c} \cup x_{c}'$ are unique. Then, by construction: $\phi(x_{c})\cdot\phi(x_{c}') = k|x_{c} \cap x_{c}'|$. However, in general, if some of the hashes are non-unique then we may either over or under-count the intersection. More formally,:
\[
    k|x_{c} \cap x_{c}'| - \Delta \leq \phi(x_{c})\cdot\phi(x_{c}') \leq k|x_{c}\cap x_{c}'| + \Delta',
\]
where $\Delta,\Delta' \leq k|x_{c} \cup x_{c}'| - Z$. That is, the error term simply counts the number of non-unique hashes amongst the symbols in $x_{c} \cup x_{c}'$.

It now remains to bound this quantity. Noting that $s \leq |x_{c} \cup x_{c}'| \leq 2s$, we may apply Lemma \ref{lemma:collisions} to conclude that, with probability at least $1-\delta$:
\[
    Z \geq ks - \frac{s^{2}k^{2}}{2d} - \max\left\{ \sqrt{\frac{2s^{3}k^{2}}{d}\log\frac{m}{\delta}}, \frac{4s}{3}\log\frac{m}{\delta} \right\},
\]
and therefore:
\[
    \Delta,\Delta' \leq \frac{s^{2}k^{2}}{2d} + \max\left\{ \sqrt{\frac{2s^{3}k^{2}}{d}\log\frac{m}{\delta}}, \frac{4s}{3}\log\frac{m}{\delta} \right\}.
\]
Putting everything together, we conclude that, with probability at least $1-\delta$:
\[
    \frac{1}{k}\phi(x_{c})\cdot\phi(x_{c}') \leq |x_{c} \cap x_{c}'| + \frac{s^{2}k}{2d} + \max\left\{ \sqrt{\frac{2s^{3}}{d}\log\frac{m}{\delta}}, \frac{4s}{3k}\log\frac{m}{\delta} \right\}.
\]
The lower bound is analogous.
\end{proof}

\end{document}